\documentclass[journal]{IEEEtran}
\usepackage{amsmath,amsfonts}
\usepackage{algorithmic}
\pdfoutput=1
\usepackage{algorithm}
\usepackage{array}
\usepackage[caption=false,font=normalsize,labelfont=sf,textfont=sf]{subfig}
\usepackage{textcomp}
\usepackage{stfloats}
\usepackage{url}
\usepackage{verbatim}
\usepackage{graphicx}
\usepackage{subfig}
\usepackage{cite}
\usepackage{float}
\usepackage[export]{adjustbox} 
\usepackage{siunitx}
\usepackage{multirow}
\usepackage{threeparttable}
\usepackage{color}
\usepackage{amsmath, bm}
\usepackage{bbding}
\usepackage[backref]{hyperref}
\usepackage{pifont}
\usepackage[all]{hypcap}
\usepackage{makecell}
\usepackage{lineno}
\usepackage{arydshln}

\begin{document}

\title{Hi-ResNet: Edge Detail Enhancement for High-Resolution Remote Sensing Segmentation}

\author{\IEEEauthorblockN{Yuxia Chen\IEEEauthorrefmark{1},
                        Pengcheng Fang\IEEEauthorrefmark{1},
                        Xiaoling Zhong\IEEEauthorrefmark{2},
                        Jianhui Yu,
                        Xiaoming Zhang,
                        Tianrui Li,~\IEEEmembership{Senior Member,~IEEE}}

\thanks{Y. Chen and X. Zhong are with the School of Mechanical and Electrical Engineering, the Chengdu University of Technology, Chengdu 610000 (email:  chenyuxia@stu.cdut.edu.cn, zhongxl@cdut.edu.cn).
P. Fang is a PhD student at the University of Southampton (email: P.Fang@soton.ac.uk).
Y. Jianhui is with the School of Computer Science, University of Sydney, Sydney, Australia (e-mail: jianhui.yu@sydney.edu.au).
Z. Xiaoming and L. Tianrui are with the School of Computing and Artificial Intelligence, Southwest Jiaotong University, Chengdu 611756, China (e-mail:  zxmswjtu@163.com; trli@swjtu.edu.cn)
\\
\IEEEauthorrefmark{1} Equal Contribution.
\\
\IEEEauthorrefmark{2} Corresponding author.
}}

\markboth{Journal of \LaTeX\ Class Files,~Vol.~14, No.~8, August~2021}%
{Shell \MakeLowercase{\textit{et al.}}: A Sample Article Using IEEEtran.cls for IEEE Journals}

\maketitle

\begin{abstract}
High-resolution remote sensing (HRS) semantic segmentation extracts key objects from high-resolution coverage areas. However, objects of the same category within HRS images generally show significant differences in scale and shape across diverse geographical environments, making it difficult to fit the data distribution. Additionally, a complex background environment causes similar appearances of objects of different categories, which precipitates a substantial number of objects into misclassification as background. These issues make existing learning algorithms sub-optimal. In this work, we solve the above-mentioned problems by proposing a High-resolution remote sensing network (Hi-ResNet) with efficient network structure designs, which consists of a funnel module, a multi-branch module with stacks of information aggregation (IA) blocks, and a feature refinement module, sequentially, and class-agnostic edge aware (CEA) loss. 
Specifically, we propose a funnel module to downsample, which reduces the computational cost, and extracts high-resolution semantic information from the initial input image. Secondly, we downsample the processed feature images into multi-resolution branches incrementally to capture image features at different scales. Furthermore, with the design of the Window multi-head self-attention, SE attention, and Depth-Wise convolution, the light-efficient IA blocks are utilized to distinguish image features of the same class with variant scales and shapes. Finally, our feature refinement module integrates the CEA loss function, which disambiguates inter-class objects with similar shapes and increases the data distribution distance for correct predictions. With effective pre-training strategies, we demonstrate the superiority of Hi-ResNet over the existing prevalent methods on three HRS segmentation benchmarks.

\end{abstract}


\section{Introduction}
\IEEEPARstart{I}{n} the geomatics community, the advancement of imaging technology allows us to obtain an increasing number of high-resolution remote sensing (HRS) images in real-time. These HRS images can be partitioned into distinct regions through pixel-level semantic segmentation, thereby providing more delicated details and features for applications such as urban planning~\cite{alshehhi2017simultaneous, gao2018building}, environmental monitoring~\cite{qin2021multilayer}, and disaster management~\cite{cooner2016detection},~\cite{xiong2020automated}. Traditional segmentation methods typically use edge-based segmentation~\cite{wanto2021combination, tian2021sobel}, threshold-based segmentation~\cite{rogerson2002change,yang2017region}, and region-based segmentation~\cite{wang2010automatic},~\cite{zhang2015segmentation} to extract key information from HRS images. However, with the rapid development of remote sensing technology, traditional methods have gradually become insufficient for complex and diverse image segmentation tasks. Consequently, in order to achieve high-precision segmentation results, many researchers opt to apply street view semantic segmentation algorithms based on convolutional neural networks~\cite{zhao2017pyramid, fu2019dual, zheng2020parsing, zhou2018unet++} and Transformer~\cite{zhuang2019shelfnet, xie2021segformer, li2019dabnet} to HRS segmentation tasks. However, these methods often perform poorly on HRS images. We attribute this to two primary reasons.

\begin{figure}[t]
    \centering
    \includegraphics[width=\linewidth]{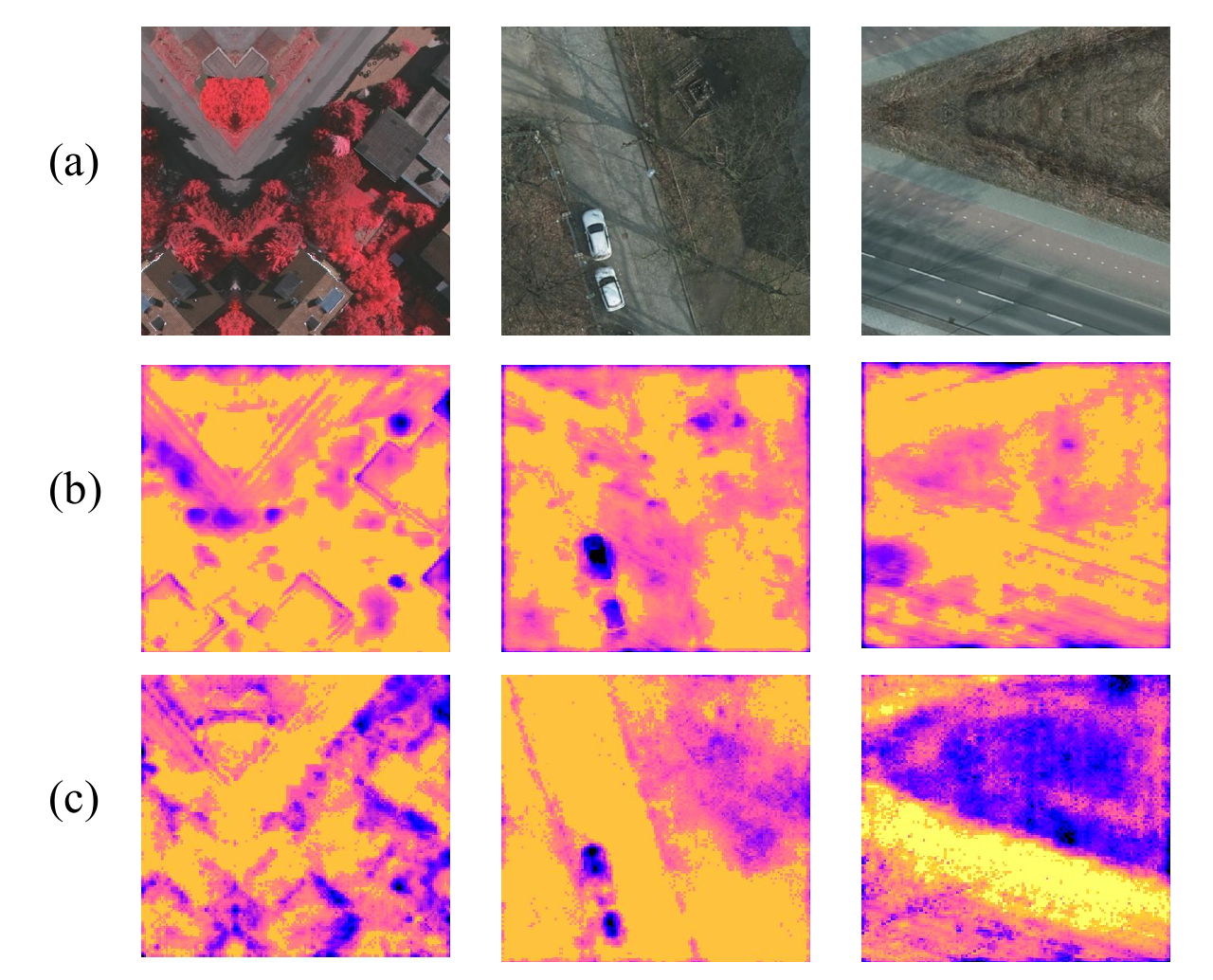}
    \caption{Comparisons of our model behavior by heatmaps with different images illustrate the feature information obtained by upsampling and merging at the end of each layer for baseline and Hi-ResNet base. The three rows (a)(b)(c) show the original image, and the features of baseline and Hi-ResNet base separately. It is evident from the results that compared to the baseline, the Hi-ResNet base extracts richer and superior feature information.}
    \label{fig:features}
\end{figure}

First, unlike conventional street-level images, objects of the same category in HRS images are often located in different geographical landscapes, leading to more scale, shape, and distribution variations for these objects~\cite{kemker2018algorithms, volpi2015semantic}. For instance, rural environments typically consist of large expanses of tree clusters and relatively narrow roads and rivers, while urban contain orderly arranged trees and wider roads and rivers~\cite{boguszewski2021landcover,marcos2018land}. Therefore, the abilities to obtain multi-scale image features and distinguish different shapes of the same objects are crucial for HRS image segmentation networks. 

Another reason is that, due to the complex background of the HRS images, objects belonging to different categories can have a similar appearance, such as flowing streams and narrow roads. Although these diverse complex scenes contribute to richer details, inter-class similarities can easily lead to model error segmentation, and severely impact the performance of semantic segmentation networks~\cite{zheng2020foreground, bai2021object}.

To address the first mentioned issue of scale variation in HRS images, some work~\cite{yin2020optimised,li2020novel,wang2023adaptive} increases the size of the receptive field by introducing the adaptive spatial pooling module, thus capturing features at different scales. Unfortunately, performing a one-to-two feature aggregation at the end of each block often loses spatial information about the features.~\cite{zhang2019multi} obtains the feature maps of the low, medium, and high scales in the first convolution of the network, and forms the dense connection modules along the diagonal. However, this approach of consecutive downsampling may lead to feature loss, and the process of dense connection also has the possibility of network structure redundancy and information blocking. In contrast to the aforementioned methods, this paper proposes the funnel module and multi-branch module. The original image passes through the inverted bottleneck (IB) block in the funnel module to obtain reliable high-resolution information. In the multi-branch module, new scale information is obtained by gradual subsampling, and features at different scales are extracted in parallel, forming an efficient and direct feature extraction convolutional stream. At the end of each feature extraction, the feature information from the previous branch was fused with the newly generated branch information. Through multi-scale information interaction,the entire network is able to obtain sufficient complete and reliable low-resolution information while maintaining high resolution. Furthermore, due to the parallel architecture of Hi-ResNet is similar to HRNet~\cite{wang2020deep}, we visualize the feature maps extracted from both the baseline (which shares the same architecture as HRNet) and Hi-ResNet base to illustrate their differences. The results are shown in Figure \ref{fig:features}. Note that there are only architectural differences between Hi-ResNet base and the baseline, which use exactly the same base blocks stacked by two stride-1, 3$\times$3 convolutions. In comparison to the baseline, our proposed model eliminates the superfluous fourth stage, while simultaneously increasing the depth of the third stage architecture. Obviously, the image features extracted by Hi-ResNet base are far beyond the image features extracted by baseline.

At the same time, in order to alleviate the issue of class distribution disparity, a common approach is to incorporate attention mechanisms into the network. For instance,~\cite{woo2018cbam,chen2021remote,chen2022gcsanet} utilize spatial attention to optimize class weights and address class imbalance problems. Additionally,~\cite{bi2021local} and~\cite{zhao2020residual} employ parallel channel attention and spatial attention to enhance local features simultaneously. On the other hand, with the proposal of vision transformer (ViT)~\cite{dosovitskiy2020image}, many subsequent works choose to apply Multi-Head Self-Attention (MHSA) in CNN-based networks~\cite{touvron2021training,wang2021pyramid,liu2022swin}. However, MHSA demands substantial computational resources when the resolution and channels number of the input feature map are large. Several studies attempt to address this issue by using local window attention~\cite{liu2021swin} or reducing the input feature resolution~\cite{wang2021pyramid,wu2021cvt}. However, for HRS images, using such modules remains a challenge.
Therefore, we present a more lightweight and efficient information aggregation (IA) block. This base block uses window-based multi-head self-attention, performing sliding operations on the channel dimension of the feature graph to capture global contextual information. At the same time, it also applies Squeeze-and-excitation (SE) attention~\cite{hu2018squeeze} to provide richer location information for the network. The IA block integrates the advantages of both convolution and MHSA, allowing it to aggregate different shapes of the same class, thereby reducing the intra-class distribution distance. Meanwhile, The use of depth-wise separable convolution in IA block reduces the parameter count of Hi-ResNet by fifty percent.

To mitigate the second mentioned of increased error segmentation due to diversification of background in HRS images and to enhance object boundary information,~\cite{li2024boundary} proposes a dual-stream network (one network for segmentation and another for boundary enhancement), designing an independent network to extract boundary information and improve segmentation results. In contrast, we propose a feature refinement module and a class-agnostic edge-aware (CEA) loss, which focuses on module and loss level. Feature refinement module upsamples the three feature maps with different resolutions obtained by muti-branch module to the same size, and concats them into a feature map. By a simple classification convolution and object-contextual representations (OCR)~\cite{yuan2019segmentation}, we obtain the results of coarse and refined segmentation of Hi-ResNet respectively and then compute them into the loss function. The outcomes of the loss function will be mixed with a ratio of 1:1. For the design of the loss functions in the HRS task, some work~\cite{zhou2018d,lin2020road} employs dice loss in road extraction by increasing the weights of the key road regions, FactsegNet~\cite{ma2021factseg} utilizes collaborative probability loss to merge the outputs of the dual-branch decoders at the probability level, aiming to enhance the utilization of information. Unlike the above losses, the proposed CEA loss in this paper focuses more on the edge information of class objects. The CEA loss expands the original Hausdorff distance (HD) loss~\cite{karimi2019reducing} to multi-classes and reduces computing resource consumption. This correction of CEA at the edge level improves the model’s perception of boundaries and shapes, enhancing its ability to capture accurate object edges. Finally, we evaluate the proposed method on widely used datasets. This study contributes four main points:

\begin{itemize}
\item[(1)] We propose the funnel module to reduce computing costs, efficiently extract high-resolution information, and avoid feature loss from the input image.
\item[(2)] We apply our proposed IA block to a multi-branch module, integrating the dynamic global modeling capability of the Transformer into CNN-based networks.
\item[(3)] We develop the CEA loss, which emphasizes edge information while taking into account multiple classes.
\item[(4)] Our Hi-ResNet is validated on several benchmarks with performance better than existing prevalent methods.
\end{itemize}

The paper is organized as follows: Section II provides an overview of related work, including Semantic Segmentation in Remote Sensing, Attention Mechanisms and Model Pre-training. In Section III, we describe the proposed method, which includes the Hi-ResNet model, the design of loss functions and the use of unsupervised and supervised pre-training in HRS tasks. Section IV presents a series of ablation experiments, and experimental results and analyses on different datasets. Finally, in Section V, we conclude the paper and provide a summary.

\section{RELATED WORK}
\subsection{Semantic Segmentation in Remote Sensing}
Parallel multi-resolution architectures primarily focus on high-level semantic information, resulting in a semantically richer and spatially more accurate representation, providing an advanced technical reference for HRS semantic segmentation tasks. Among them, HRNet~\cite{wang2020deep} was well known as a parallel semantic segmentation model that could maintain high resolution. It passes through four stages of gradually decreasing resolution and performs multi-scale fusion to enhance high-resolution representations. Subsequently, researchers attempted to combine HRNet with object-contextual representations~\cite{yuan2019segmentation} which distinguishes contextual information for the same target category from different target categories and optimizes feature pixels. This architecture was widely applied in the field of HRS segmentation~\cite{cheng2020remote}. However, HRNet primarily focused on high-resolution semantic features of images, while object-contextual representations was more concerned with the relationships between image objects and their pixels, both of which ignore the high-level semantic information. Unfortunately, reliable high-level semantic information that includes target locations is undoubtedly crucial, as HRS images often contain a large amount of complex and unrelated background, and small target objects usually occupy only a few pixels. To obtain richer high-level semantic information, some studies~\cite{hamaguchi2018effective,chen2018encoder,kirillov2019panoptic,zhou2018unet++} applied different dilated convolutions to multiple features of traditional CNN networks. By expanding the receptive field of the convolution kernel, these studies have constructed distinctive local semantic representation modules, thereby effectively utilizing multi-scale features. Furthermore, some researchers~\cite{cai2020remote,xu2021deep} applied graph convolution on multi-layer features, treating each pixel as a node, and then connecting the extracted graph features with the final global visual features. Despite this, locally aggregating features in the spatial direction might overlook channel and positional information of high-level semantics. An effective solution is to establish an information connection between space and channels in convolutional networks. Recently, MBFANet~\cite{shi2023remote} combined the pooling channel attention module and convolutional coordinate attention module to complement each other, which helped the models focus on more complex background categories. 
SAPNet~\cite{zheng2022sapnet} joint models both spatial and channel affinity, which allows for preserving spatial details and extracting accurate channel information. 
Inspired by the parallel architecture of HRNet, we propose Hi-ResNet in this work. The Hi-ResNet utilizes a funnel module and muti-branch module to obtain rich high-resolution semantic information and use feature refinement module to enhance small target features.

\begin{figure*}[t]
    \centering
    \includegraphics[width=\linewidth]{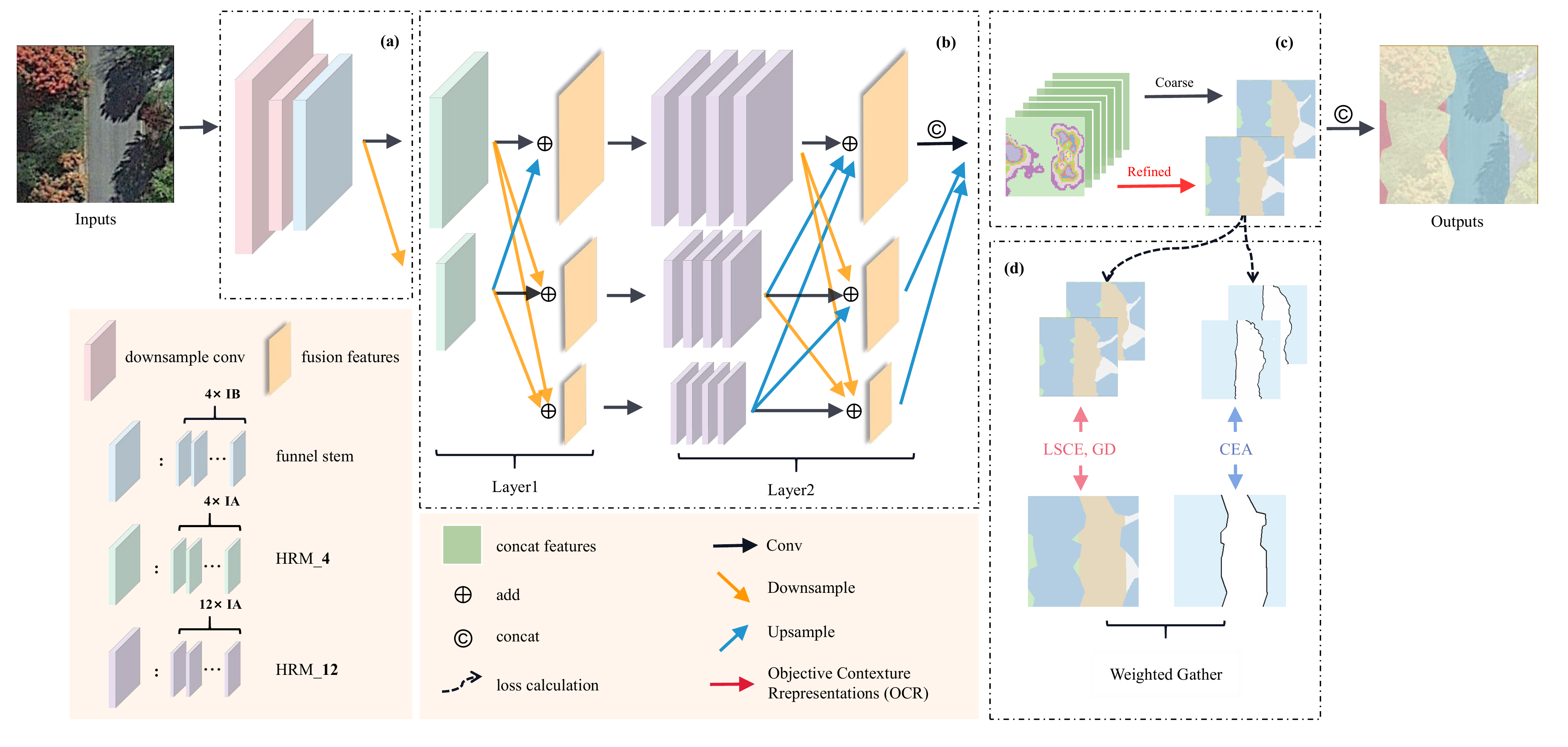}
    \caption{The comprehensive architecture of Hi-ResNet is partitioned into four components. (a) The funnel module, composed of a downsample part and a funnel stem, is proposed for downsampling input imagery and facilitating feature extraction. (b) The multi-branch module further hones these features via the amalgamation of a multi-resolution convolutions stream. (c) In the feature refinement module, coarse features are computed directly via a convolution layer, with refined features managed through the utilization of OCR~\cite{yuan2019segmentation}. During inference, the coarse results and refined results are added in a 1:1 ratio as the model's output. (d) Multiple loss functions are employed, including LSCE loss~\cite{muller2019does} and GD loss\cite{sudre2017generalised}, which are computed in direct relation to the ground truth and predictions. Concurrently, the CEA randomly elects a category, designating all others as background, computing the loss between the two categories.}
    \label{fig:Hi-ResNet}
\end{figure*}

\subsection{Attention Mechanisms}
Attention mechanisms could help the network to locate the information of interest and inhibit useless information, which has been widely used in convolutional neural networks~\cite{hu2018squeeze,fu2019dual,liu2020improving,li2021abcnet,yamazaki2023aerialformer}. For HRS tasks, some studies utilized the popular attention mechanism Squeeze-and-Excitation to automatically process the features of various scenes and extract more effective features~\cite{tian2021semsdnet,zhang2022mrse,zhang2022transformer,ding2020lanet,lin2020road}. However, this attention mechanism only focused on inter-channel information while neglecting spatial and positional information about the features. To simultaneously capture channel and positional information, researchers explored the use of Convolutional Block Attention Module or Bottleneck Attention Module in network architectures~\cite{wang2022cbam,zhu2021improving,shi2021deeply}. These modules used spatial attention to obtain the location information and reduce the input channel dimension to save the calculation cost. Due to the limited receptive field of the sliding window in convolutional operations, only local relationships were captured, it could not maintain long-range dependencies between different positions in the image.

Currently, some works apply the Transformer to semantic segmentation models in HRS, thereby obtaining global information~\cite{strudel2021segmenter,xie2021segformer,xu2023rssformer}. DC-Swin~\cite{wang2022novel} introduced the Swin-Transformer for the encoder in fine image segmentation, while Unetformer~\cite{wang2022unetformer} uses Unet as an encoder, proposing a global-local attention mechanism to construct Transformer blocks in the decoder. Nevertheless, the substantial computational complexity introduced by self-attention made the training cost of the network expensive. This poses challenge for its application in lightweight convolutional networks. To transfer the dynamic global modeling capability of the Transformer to CNN-based networks while keeping the network lightweight, we propose the efficient IA block. This block combines the long-range interaction capability of the Transformer with the inductive bias of CNNs. By using window-based multi-head self-attention and SE attention, it provides accurate feature information for the model. Meanwhile, the use of depth-wise separable convolution brings less computational quantities and parameters for IA block.

\subsection{Model Pre-training}
In addition to the design of the network itself, excellent pre-training programs are also indispensable. Numerous studies showed that applying pre-training can make models more stable and extract more commonalities~\cite{cha2023billion,wang2022advancing,wang2022empirical,ayush2021geography}. Therefore, we pre-train the Hi-ResNet to enhance the fine-tuning ability for HRS segmentation tasks. Recently, for HRS tasks, some studies~\cite{workman2023handling,kong2020enhanced} used labeled semantic segmentation datasets such as Mapillary~\cite{neuhold2017mapillary} for pre-training to improve model performance. However, these large-scale labeled datasets mostly come from natural images, and pre-training on them for HRS tasks often yields poor results. It is worth noting that recent works on unsupervised pre-training~\cite{khan2022transformers,chen2020simple,chen2020improved,he2020momentum} showed that unsupervised pre-training outperforms the supervised way in downstream tasks such as segmentation.
MoCo~\cite{he2020momentum}, as a mechanism for building dynamic dictionaries for contrastive learning, surpassed its supervised counterpart in seven downstream tasks.~\cite{zhao2020makes} illustrated that MoCo mainly transferred low-level and middle-level semantic features, and when performing image reconstruction, the reconstructed images without supervision were closer to the original data distribution. Based on the previous work, we argue that employing supervised pre-training will provide richer and more comprehensive prior information for HRS tasks. At the same time, using the pre-training mechanism of MoCo can effectively compensate for the loss of precise localization information in the network and reduce the emphasis on local object information. Therefore, in this study, we apply both fully supervised and unsupervised pre-training strategies on Hi-ResNet and evaluate the performance of the two methods.

\section{PROPOSED METHODS}
In this section, we present the framework of Hi-ResNet, including the funnel module, the multi-branch module with information aggregation blocks, and feature refinement module. Then we introduce the class-agnostic edge aware loss for HRS image feature extraction. Finally, we present how to transfer the various pre-training strategies to the HRS segmentation task.

\subsection{Hi-ResNet Framework}
The Hi-ResNet proposed in this paper is shown in Figure \ref{fig:Hi-ResNet}. In the following sections, we will present the funnel module, multi-branch module, and feature refinement module, and the implementation details of each module in turn.

\subsubsection{Funnel Module}
\begin{figure}[!ht]
    \centering
    \includegraphics[width=\linewidth]{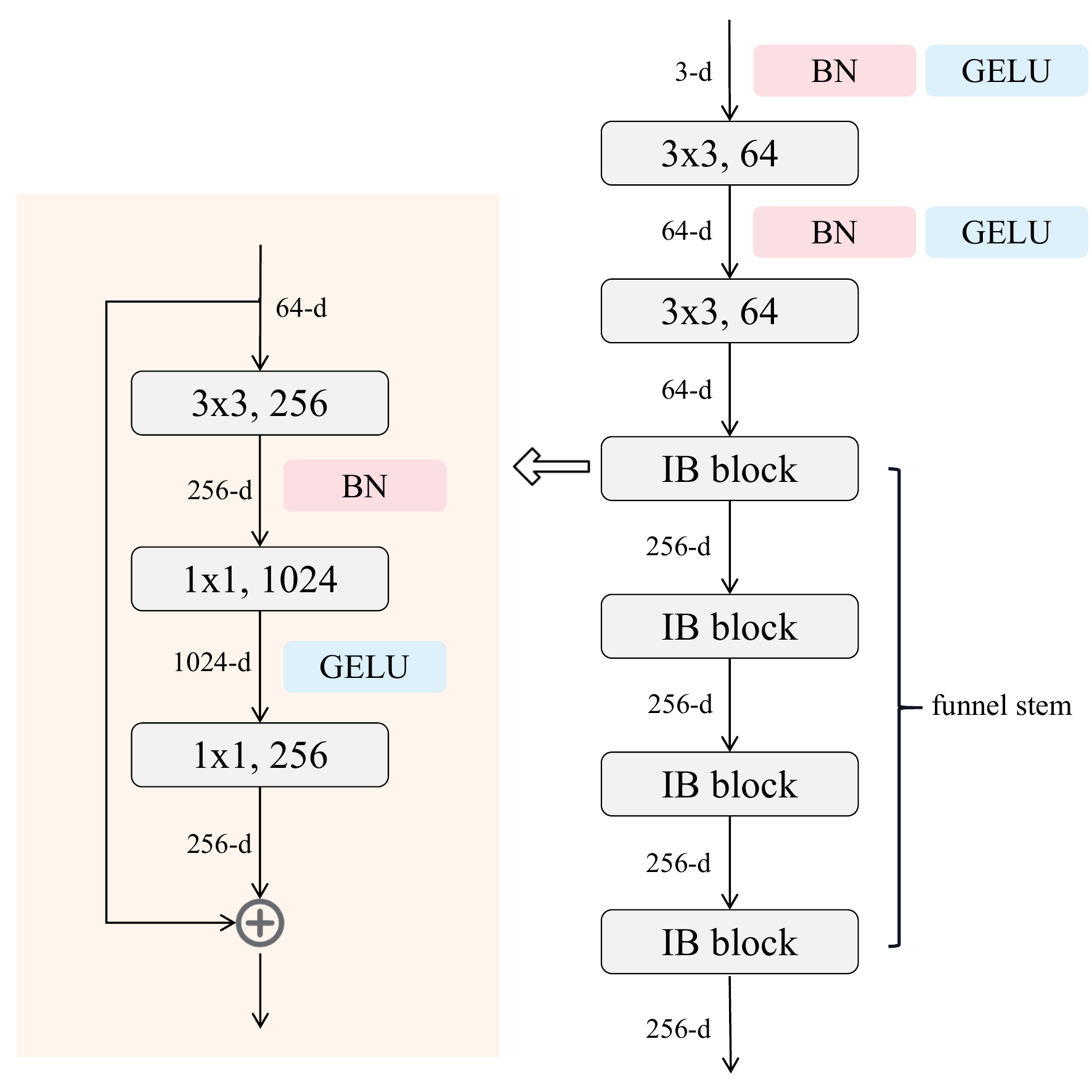}
    \caption{The structure of the funnel module where IB refers to inverted bottleneck. The number in each block refers to the kernel size and channel numbers respectively.}
    \label{fig:funnel}
\end{figure}

In the funnel module, we start by passing the input image through two stride-2, 3$\times$3 convolutions, which reduce the image resolution to 1/4 of its original size. During the network downsampling, the batch normalization (BN) layer was placed before the convolution operation. It could improve the generalization and stability of the model by applying the BN layer, which makes the pre- and post- samples to different Gaussian distributions. Then, the image goes through a funnel stem with four inverted bottleneck (IB) blocks to obtain high-resolution semantic features. The traditional bottleneck block uses a structure with long heads and a short middle. With consideration of the distribution characteristics of HRS data, to prevent the collapse of activation space and loss of channel information caused by non-linear activation functions in network layers~\cite{sandler2018mobilenetv2}, our work adapts the IB block with thin heads and a thick middle.
This block is used to extract richer semantic features by performing high-dimensional upsampling on HRS images, followed by residual connection and linear activation function to avoid information loss, thereby preserving more complete information of HRS images. The design of the funnel module is illustrated in Figure \ref{fig:funnel}. For IB block, we use a stride-1, 3$\times$3 convolution in the first layer. In the middle layer, we apply a stride-1, 1$\times$1 convolution to quadruple the number of channels, thus obtaining richer high-resolution semantic information. Subsequently, a stride-1, 1$\times$1 convolution is used in the final layer to revert the channel count to its original number.

\subsubsection{Multi-branch Module}
The multi-branch module consists of multi-resolution convolution streams and repeated feature fusions. First of all, to address the issue of unstable segmentation accuracy caused by differences in image scales, our network maintains the high-resolution representation of the input image throughout generating a new low-resolution branch at each end of the layer.

\begin{figure}[!ht] 
    \centering
    \includegraphics[width=\linewidth]{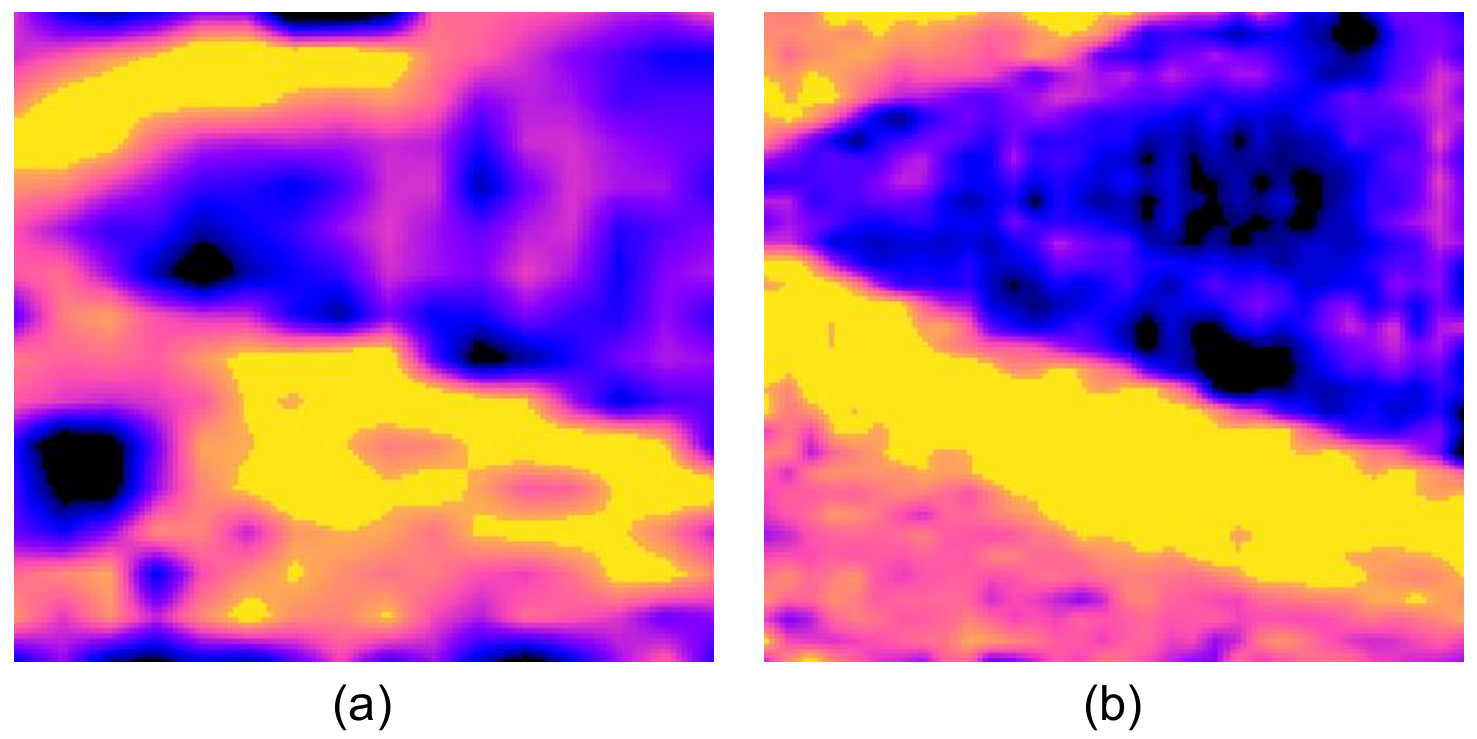}
    \caption{(a) the output features of the multi-branch module in Hi-ResNet before the extension. (b) the output features of the multi-branch module in Hi-ResNet after the extension.}
    \label{fig:features2}
\end{figure}

We employ the parallel approach to conduct a series of convolution operations in multi-resolution branches, forming the multi-resolution convolution stream. Notably, the minimum resolution of the image in the parallel branch of the second layer is only 1/16 of the original image, indicating that this layer focuses more on the high-level semantic information of the image. Due to the significant layout differences among objects in different areas of HRS images, the shape and contour features in high-level semantic information are crucial. This paper argues that it cannot extract rich high-level semantic features that contain target locations if merely stacking the same number of blocks as in other layers and using the same sliding window sampling. Therefore, we stack 4 IA blocks as high-resolution module\_4 (HRM\_4) in the first layer, and triple the number of IA blocks in the second layer, i.e., 12 IA blocks as high-resolution module\_12 (HRM\_12). Figure \ref{fig:features2} illustrates the semantic information extracted by the multi-branch module before and after the extension. More abundant and reliable high-level semantic information can be obtained through the second layer after expansion, which not only effectively alleviates the problem of class distribution inconsistency and reduces intra-class variance but also avoids the loss of positional information of small target objects that occupy only a few pixels in the image, thereby enhancing the weak features of small target objects.

\begin{figure}[!ht]
    \centering
    \includegraphics[width=\linewidth]{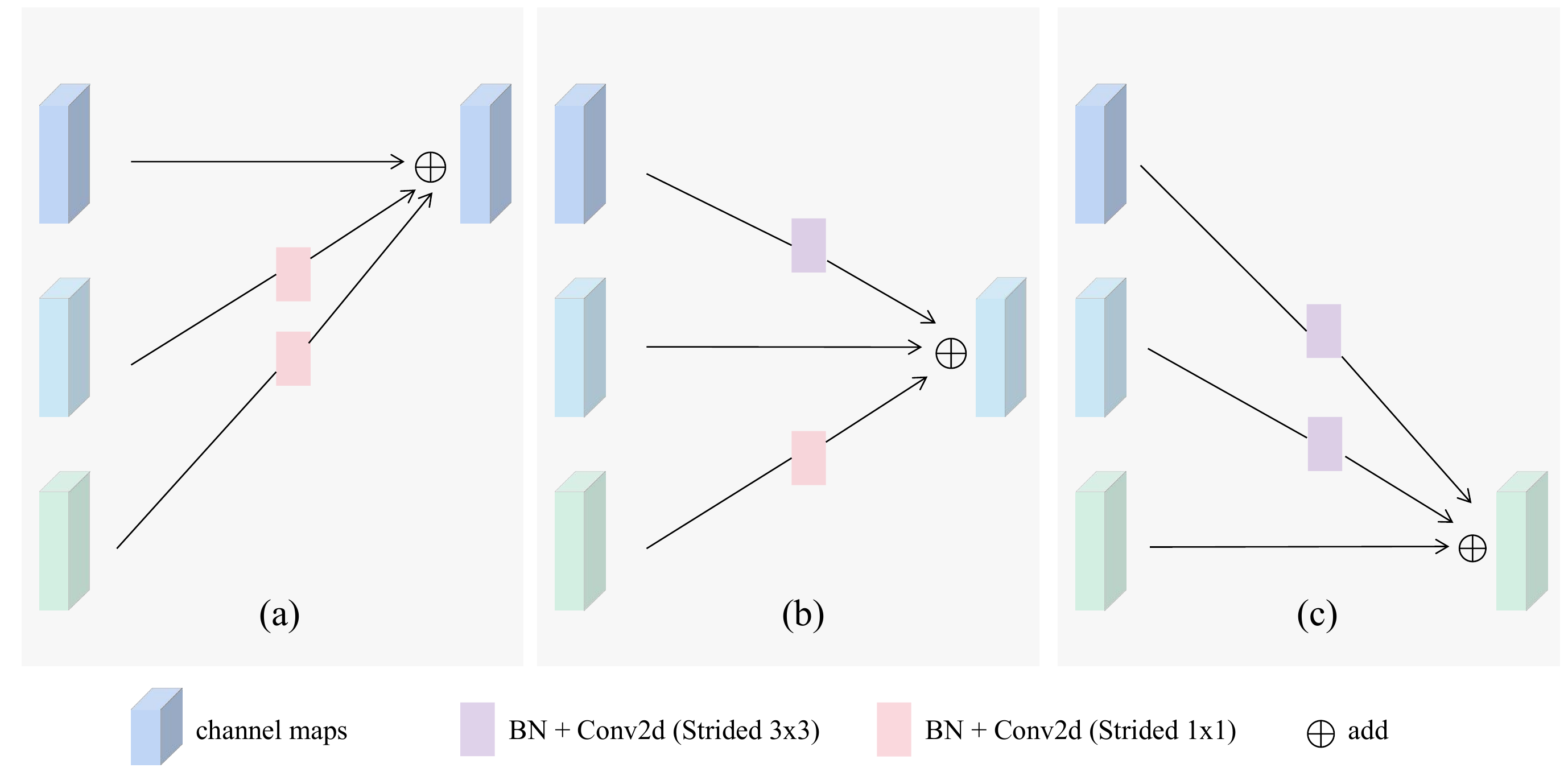}
    \caption{This figure illustrates the process of feature information aggregation across various resolutions in the fusion layer of the network. Furthermore, we exchange the sequence of the BN and the conv here.}
    \label{fig:fusion}
\end{figure}

Numerous studies propose methods for multi-scale feature fusion~\cite{long2015fully,zhao2017pyramid,chen2018encoder}. Classic semantic segmentation networks like UNET~\cite{ronneberger2015u} and SegNet~\cite{badrinarayanan2017segnet} extract feature maps of different resolutions during the downsampling phase. In the model's upsampling phase, these are combined with feature maps of the corresponding resolution, serving to prevent feature loss. In contrast, our approach performs cross-layer fusion between parallel branches with different resolutions, capturing features of different sizes by repeatedly exchanging information on different scales at each layer. Figure \ref{fig:fusion} illustrates the fusion process for layer2, where the input consists of three images with different resolutions. Different sampling methods are used depending on the resolution of the input and output. The upsampling stage includes bilinear upsampling, BN layer, and a stride-1, 1$\times$1 convolution, while the downsampling stage includes BN layer and a stride-2, 3$\times$3 convolution. We sum the images sampled at the same resolution to produce the final output for that resolution. The multi-branch module process ultimately outputs three feature maps with different resolutions.

\subsubsection{Information Aggregation Block}
\begin{figure}[!ht]
    \centering
    \includegraphics[width=\linewidth]{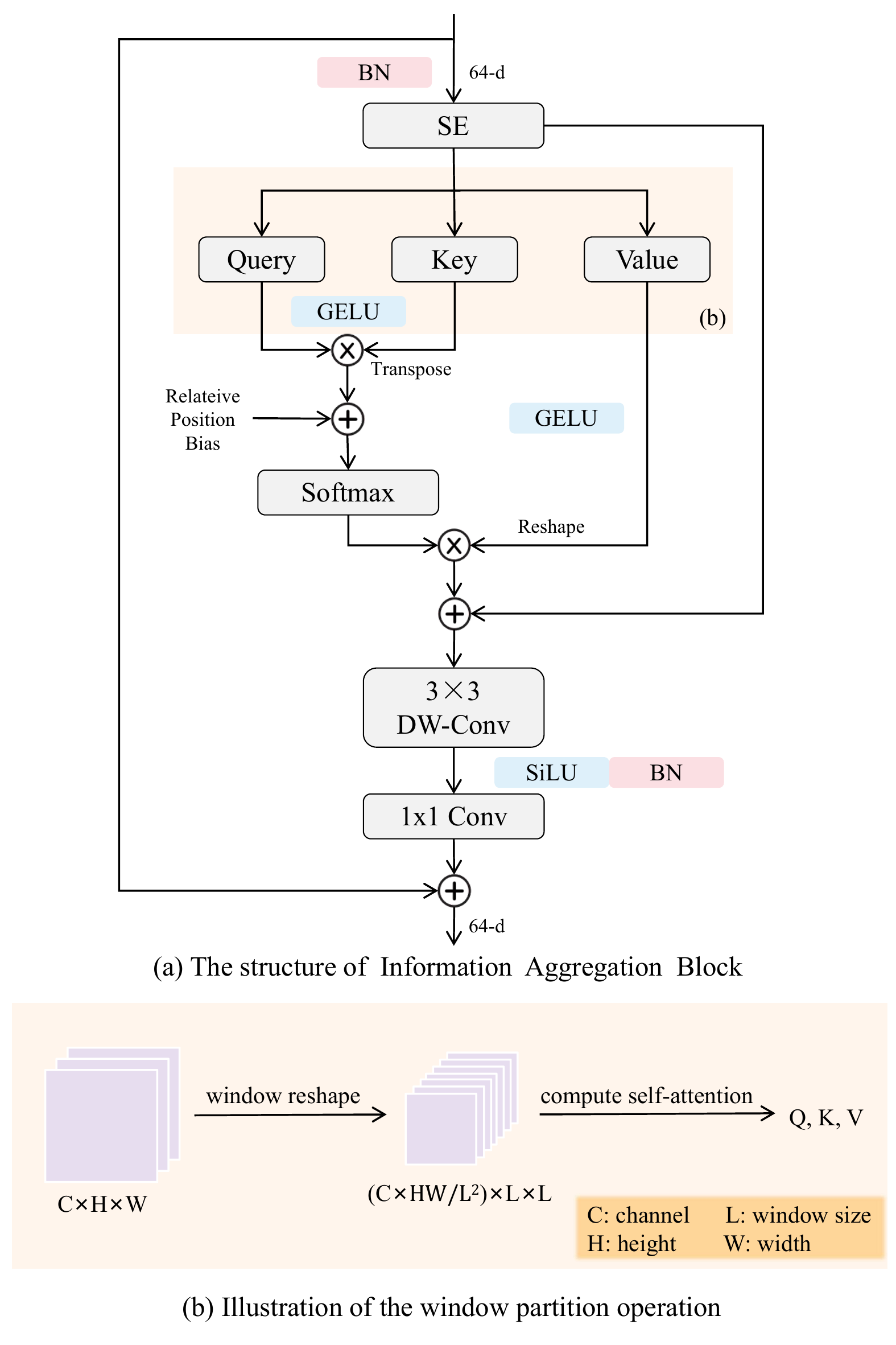}
    \caption{The IA block mainly consists of two attentions: a convolutional self-attention and a SE attention, a Depth-Wise convolution with a relatively large kernel, and a 1$\times$1 convolution.}  
    \label{fig:IA}
\end{figure}

HRS images provide rich details and features but also bring more irrelevant background objects. To suppress the impacts brought by the irrelevant background information and to enhance the spatial and positional feature representations, we propose a lightweight block: Information Aggregation (IA) block. This block refers to few parameters, easy operators, and high efficiencies. In IA block, the SE attention~\cite{hu2018squeeze} is utilized to sufficiently sketch the hidden information from the input features. Then, we consider that the attention mechanism shares a wider perception field than the convolutional kernel, thereby multi-head self-attention (MHSA) applied in the block. However, MHSA consumes huge computation resources, especially in image calculations. Therefore, Window-MHSA (WMHSA) and Depth-Wise Convolution (DW-Conv) both with a skip connection are utilized to trade-off model cost and accuracy. Unlike sliding windows of the Swin-Transformer~\cite{liu2021swin}, the WMHSA here simply resize the tensors from $C\times H\times W$ to $(C\times H\times W/L^{2})\times L\times L$ (shown in Figure~\ref{fig:IA}(b)) and then conduct MHSA. We employ another skip connection acting on the whole block, which enables feature reuse and prevents loss. In the IA block, instead of RELU, we prefer GELU and SILU to obtain a relatively slight change during the minus. Moreover, different from general convolutional blocks, we utilize fewer activation functions and normalizations. The IA block is illustrated in Figure \ref{fig:IA}.

\subsubsection{Feature Refinement Module}
We use the three different resolution feature maps output by the multi-branch module as inputs to the feature refinement module. In the feature refinement module, we combine the three input images to the same size using bilinear upsampling, which serves as the coarse segmentation of the network. By leveraging OCR~\cite{yuan2019segmentation}, we first treat a category in the coarse segmentation result as a region and estimate the comprehensive feature representation within that region by aggregating the representations of each pixel. Then, we compute the pixel-region relationships to obtain corresponding weights, which are used to enhance the representation of each pixel by weighting all the regions. The weighted feature representation serves as the refined segmentation result of the model. Lastly, Hi-ResNet outputs both coarse segmentation and refined segmentation. 

\begin{table}[!ht]
  \renewcommand{\arraystretch}{1.8}
  \begin{center}
  \footnotesize
    \caption{THE MAIN ARCHITECTURE CONFIGURATION OF Hi-ResNet}
    \resizebox{\columnwidth}{!}{
    \begin{tabular}{c|c | c |c}
    \hline
        \multirow{2}{*}{Input Size} &  \multirow{2}{*}{Funnel Module} & \multicolumn{2}{c}{Multi-branch Module} \\
        \cline{3-4}
        & &  Layer1 & Layer2\\
        \hline
       $H \times W \times 3$ & \makecell[c]{$[3\times3, stride=2]\times2$ \\ $[IB~Block]\times B_{1}$} &  & \\
       \hline
       $\frac{4}{H} \times \frac{4}{W} \times C_{1}$ &  & $[IA~Block]\times B_{2} \times M_{1}$ & $[IA~Block]\times B_{3} \times M_{2}$\\
       \hline
       $\frac{8}{H} \times \frac{8}{W} \times C_{2}$ &  &   $[IA~Block]\times B_{2} \times M_{1}$& 
       $[IA~Block]\times B_{3} \times M_{2}$\\
       \hline
       $\frac{16}{H} \times \frac{16}{W} \times C_{3}$ & & &$[IA~Block]\times B_{3} \times M_{2}$\\
      \hline
    \end{tabular}
    }
  \end{center}
  \label{table:1-1}
\end{table}

\begin{table}[!ht]
  \renewcommand{\arraystretch}{1.5}
  \begin{center}
  \scriptsize
    \caption{THE CONFIGURATIONS OF Hi-ResNet INSTANCES}
    \resizebox{\columnwidth}{!}{
    \begin{tabular}{c| c | c | c }
    \hline
       Model & \makecell[c]{Channels \\ $(C_{1}, C_{2}, C_{3})$} & \makecell[c]{Blocks \\ $(B_{1}, B_{2}, B_{3})$} & \makecell[c]{Modules \\ $(M_{1}, M_{2})$} \\
      \hline
       Hi-ResNet & (48, 96, 192) &  (4, 4, 12)& (1, 4) \\
      \hline
    \end{tabular}
    }
  \end{center}
  \label{table:1-2}
\end{table}

The main architecture configuration of Hi-ResNet is shown in Table \ref{table:1-1}. Here, H and W denote the height and width of the input image, respectively. C1, C2, and C3 denote the number of channels. B1, B2, and B3 denote the number of blocks on each branch of the module, respectively. M1 and M2 denote the number of modules in each layer, respectively.
Table \ref{table:1-2} displays the detailed configuration of (C1, C2, C3), (B1, B2, B3) and (M1, M2) instances in Hi-ResNet.

\subsection{Loss Design}
In HRS tasks, semantic segmentation typically involves more than two labels, with significant differences in the number and pixel range of objects for different categories, leading to sample imbalance and sub-optimal performance. Therefore, the appropriate loss function is crucial. In this work, we propose a new class-agnostic edge aware (CEA) loss, which is combined with the Generalised Dice loss (GD)~\cite{sudre2017generalised} and Label Smoothing Cross-Entropy loss (LSCE)~\cite{muller2019does} as the training loss.

\subsubsection{Generalised Dice Loss}
Weighted cross-entropy and Sensitivity-Specificity approaches are designed to address imbalanced problems only in binary classification tasks. In contrast, the GD loss method can weight various pixel classes, allowing for a more comprehensive approach to imbalanced sample issues. The loss calculation for GD loss can be expressed as:
\begin{equation}
{L}_{GD} = 1 - 2\frac{\sum_{l=1}^2w_l\sum_{n}r_{ln}p_{ln}}{\sum_{l=1}^2w_l\sum_{n}r_{ln} + p_{ln}}
\end{equation}

The equation for GD loss involves using $r_{ln}$ to represent the label of each pixel in the reference foreground segmentation for class $l$, and $p_{ln}$ to denote the predicted probabilistic map for the foreground label of class $l$ over $N$ image elements $p_{n}$. The weighting factor $w_{l}$ is used to provide invariance to different label set properties. Its calculation is expressed as:
\begin{equation}
w_l = \frac{1}{\sum_{i=1}^Nr_{ln}^2}
\end{equation}

During the calculation process, overlapping $r_{n}$ and $p_{n}$ for each category $l$ are added according to their weights and then divided by the weighted sum of the union part. This effectively suppresses the interference of complex background classes, enhances the features of small targets, and alleviates the problem of imbalanced image samples.

\subsubsection{Label Smoothing Cross-Entropy Loss}
The label smooth technique proposed in~\cite{muller2019does} as a training strategy can adjust the extreme values of the loss and improve the model's generalization ability when combined with Cross-Entropy loss. The equation of Label Smoothing Cross-Entropy loss is as follows:
\begin{equation}
\mathcal{L}_{lsce} = -\sum_{k=1}^K\sum_{n=1}^Ny_k^{(n)}\log\hat{y}_k^{(n)}
\end{equation}
\begin{equation}
y_k^{(n)}=
\begin{cases}
1 - \varepsilon& \text{if } n = k,\\ 
\varepsilon/(K - 1)& \text{otherwise}
\end{cases}
\label{eq:lsce}
\end{equation}

In the above equation, $N$ and $K$ indicate pixel values and categories, respectively. $y_k^{(n)}$ and $\hat{y}_k^{(n)}$ represents the sample label following the label smoothing operation and the corresponding softmax output belonging to the category $k$, respectively. Equation~\ref{eq:lsce} shows the calculation process of $y_k^{(n)}$. When the pixel representation class $n$ is the same as the input class $k$, $y_k^{(n)}$ equals $1 - \varepsilon$, where $\epsilon$ is the smoothing factor. Otherwise, $y_k^{(n)}$ equals $\varepsilon/(K - 1)$, where $K$ is the total number of classes.

Considering that HRS image datasets usually have a small amount of data, we argue that using this loss can prevent overfitting of the network and provide the correct optimization direction for the model.

\subsubsection{Class-agnostic Edge Aware Loss}
Our proposed CEA loss enhances the original Hausdorff distance (HD)~\cite{karimi2019reducing} loss in two crucial stages. Initially, we extend the Hausdorff loss to accommodate multiple classes. Subsequently, the HD loss utilizes the Scipy library to compute the Euclidean distance transform, an approach that has proven to be inefficient in the context of multi-class loss calculation. Hence, we employ cascaded convolutional operations to approximate the Manhattan distance transform of images, thereby addressing the observed inefficiency. It is shown below:

\begin{equation}
    \mathcal{P} = Softmax(s_{\theta}(p))
\label{eq:softmax}
\end{equation}

\begin{equation}
       \mathcal{T} = Onehot(t)
    \label{eq:one_hot}
\end{equation}

\begin{equation}
    \mathcal{L}_{CEA} = \int_{\Omega}(\mathcal{T} - \mathcal{P})^2(D_{G}(\mathcal{T})^\beta + D_{S}(\mathcal{P})^\beta)\mathrm{d}\mathcal{P}
\label{eq:l_hd}
\end{equation}

We get $\mathcal{P}$ from Equation~\ref{eq:softmax} and $\mathcal{T}$ Equation~\ref{eq:one_hot}, where p refers inputs of our model $s_{\theta}$, and $t$ refers the ground truth. In Equation~\ref{eq:l_hd}, $\Omega$ denotes the spatial domain of the training images. The distance function from the predicted boundary $S$, after applying thresholds $s_{\theta}$, is represented by $D_{S}$. The hyper-parameter $\beta$, set to 2 by the authors of~\cite{karimi2019reducing} through a grid search, is also a part of this process.

\begin{figure*}[t]
    \centering
    \includegraphics[width=\linewidth]{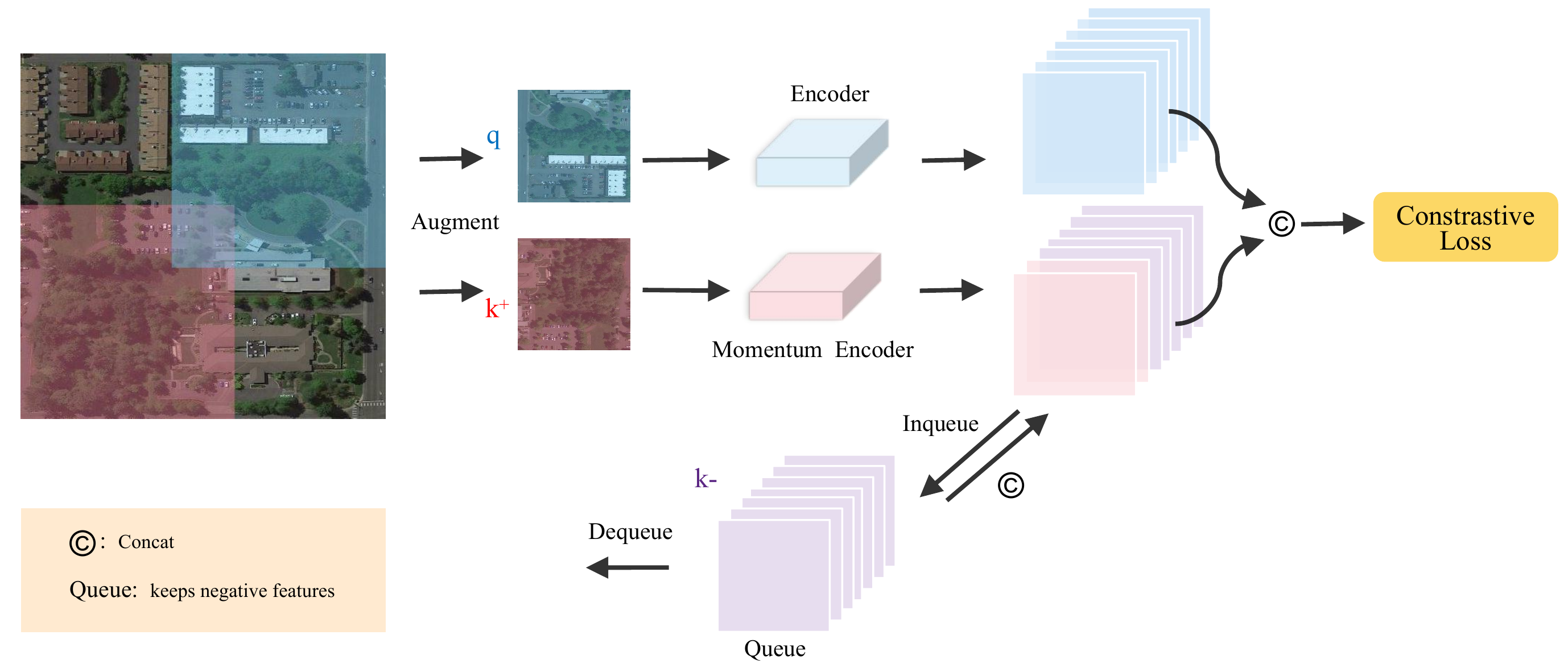}
    \caption{The figure outlines the entire process of MoCoV2 pre-training. $\bm q$ and $\bm k^{+}$ are the positive sample and the negative respectively, augmented from the same image.  while $\bm k^{-}$ refers to the past negative features stored in the queue.}
    \label{fig:moco}
\end{figure*}

\subsubsection{Loss function}
According to the results of the ablation study, the overall loss can be formulated as:
\begin{equation}
\mathcal{L}_{coarse/refined} = \alpha\mathcal{L}_{GD} + \beta\mathcal{L}_{LSCE} + \gamma\mathcal{L}_{CEA}
\label{equation:losses}
\end{equation}

\begin{equation}
\mathcal{L} = \mathcal{L}_{coarse} + \mathcal{L}_{refined}
\end{equation}

where $\mathcal{L}_{coarse}$ and $\mathcal{L}_{refined}$ represent the coarse segmentation and refined segmentation of the model, respectively. Both $\mathcal{L}_{coarse}$ and $\mathcal{L}_{refined}$ are calculated by Equation~\ref{equation:losses}, where $\alpha$, $\beta$, $\gamma$ denote the weights of each loss. In our model, they are set to 0.3923, 0.3923, 0.2153 respectively.

\subsection{Remote Sensing Pre-training}
Given that this paper introduces a new backbone, it is necessary to employ different strategies for pre-training of Hi-ResNet. In this section, we demonstrate two pre-training strategies designed for the HRS semantic segmentation task.

\subsubsection{Dataset}
\begin{itemize}
\item[$\bullet$] The Mapillary dataset~\cite{neuhold2017mapillary} is presently the largest publicly available street view dataset, with specific instance annotations and a high degree of diversity. This dataset encompasses 25,000 high-resolution RGB images, captured by a variety of imaging devices, and includes fine-grained labels for 66 categories.
\end{itemize}

\begin{itemize}
\item[$\bullet$] Million-AID~\cite{long2021creating} is a comprehensive benchmark dataset designed for remote sensing scene classification. This dataset obtains images with resolutions ranging from 0.5m to 153m from multiple satellites of Google Earth. The scene labels are obtained through the geographical coordinate information, resulting in over one million images labeled with 51 semantic scene categories.
\end{itemize}

\subsubsection{Pre-training Details} 
The section introduces two distinct methods for pre-training models. For the supervised training, the Mapillary dataset is selected as it shares the same downstream task as this paper. We got 2 million 256$\times$256 images after clipping images of Mapillary. To address the issue of imbalanced data, we filter the cropped images based on the proportion of pixel classes within each labels, ultimately resulting in 400,000 images for training. The supervised pre-training process is the same as for training our own model.

The unsupervised training utilizes the Million-AID dataset. Since the images in Million-AID have varying resolutions, they are partitioned into 400$\times$400, while images size less than 400 are dropped from the dataset. We use contrast learning MoCoV2~\cite{he2020momentum} as the unsupervised pre-training method. The process of MoCoV2 is shown in Figure \ref{fig:moco}, and the primary training settings for both pre-training approaches are presented in Table \ref{table:2}.

\begin{table}[!ht]
  \begin{center}
  \scriptsize
    \caption{HYPER-PARAMETER SETTINGS OF DIFFERENT PRE-TRAINED MODELS}
    \begin{tabular}{l|c c c c c c}
        \hline
      \textbf{Method} & \textbf{dataset} & \textbf{lr} & 
      \textbf{image size} & \textbf{batch size} & \textbf{quantity}\\
      \hline
      supervised & Mapillary & 5e-5 & 256$\times$256 & 80 & 400,000\\
      MoCoV2 & MillionAid & 0.015 & 400$\times$400 & 64 & 1,000,000\\
      \hline
    \end{tabular}
  \end{center}
  \label{table:2}
\end{table}

MoCoV2 pre-training commences with the augmentation of a batch of images twice, to generate the positive samples, denoted as $q$, and the batch negative samples, denoted as $k^{+}$. Following this, the logits of $q$ and $k^{+}$ are procured by introducing $q$ and $k^{+}$ into the standard encoder and the momentum encoder, respectively. In addition, the input negative samples are amalgamated with $k^{+}$ and $k^{-}$, which are retrieved from the queue. Subsequently, the logits of the positive and negative samples are concatenated, following which the InfoNCE loss function is computed to update the standard encoder:

\begin{equation}
    \mathcal{L}_{q, k^{+}, {k^{-}}} = -\lg \frac{\exp{(q\cdot k^{+} / \tau)}}{\exp{(q\cdot k^{+} / \tau)} + \sum\limits_{k^{-}}\exp{(q\cdot k^{-} / \tau)}}  
\label{eq:moco1}
\end{equation}
where \textbf{q} is a query representation, $\bm k^{+}$ is a representation of the positive (similar) key sample, and $\bm k^{-}$ are representations of the negative (dissimilar) key samples. $\bm \tau$ is a temperature hyper-parameter. During training, only the normal encoder updates while the momentum encoder is updated with the function below:
\begin{equation}
    \mathcal{\theta}_{k} = m\theta_{k} + (1 - m)\theta_{q}
\label{eq:moco2}
\end{equation}
Here \textbf{m} $\in [0, 1)$ is a momentum coefficient. Only the parameters $\theta_{q}$ are updated by back-propagation. The momentum update in Equation~\ref{eq:moco2} makes $\theta_{k}$ evolve more smoothly than $\theta_{q}$. Finally, $k^{+}$ will be added to the queue, and features earlier in the queue will be dequeued.

\section{EXPERIMENTAL RESULTS AND ANALYSIS}
In this section, we evaluate the performance of our proposed model on multiple remote sensing datasets, including LoveDA, Potsdam, and Vaihingen. We first conduct a series of ablation studies to analyze and identify a suitable framework for our proposed model. Next, we compare our Hi-ResNet with current state-of-the-art (SOTA) methods on public benchmarks. Additionally, we demonstrate the performance of our proposed model and existing popular frameworks in terms of computational complexity, inference speed, and memory usage on three datasets.

\subsection{Datasets}

\subsubsection{LoveDA}
The LoveDA dataset~\cite{wang2021loveda} comprises 5987 HRS images (GSD 0.3m) from three different cities, each containing 166768 annotated objects. Each image is 1024$\times$1024 pixels and includes 7 land cover categories, namely building, road, water, barren, forest, agriculture, and background. The dataset provides 2522 images for training, 1669 images for validation, and 1796 official images for testing. The images consist of two scenes, urban and rural, from three Chinese cities, namely Nanjing, Changzhou, and Wuhan. Consequently, the dataset presents a significant research challenge due to the presence of multi-scale objects, complex backgrounds, and inconsistent class distributions.

\subsubsection{Potsdam}
Potsdam is a historic city with complex buildings, narrow streets, and dense settlement structures. The Potsdam dataset is composed of 38 images, each size is 6000$\times$6000, and containing a true orthophoto (TOP) extracted from a larger TOP mosaic. The dataset has been manually classified into the six most common land cover categories (impervious surfaces, background, buildings, low vegetation, trees, and cars) , and the ground sampling distance of the TOP is 5cm. In this paper, we follow the approach used in~\cite{wang2022unetformer} and use 23 images (excluding image $7\_10$ with error annotations) for training and 14 images for Validation.

\subsubsection{Vaihingen}
The village of Vaihingen comprises many individual buildings and small multi-story houses. This dataset includes 33 HRS images, and the average size of images is 2494$\times$2064. Each image consists of three channels: near-infrared, red, green, and a single-band DSM (note that the DSM is not used in our experiments). All images are labeled into the same six classes as the Potsdam dataset. For the experiment, we follow~\cite{wang2022unetformer} to select the remote sensing images with ID 2, 4, 6, 8, 10, 12, 14, 16, 20, 22, 24, 27, 29, 31, 33, 35, and 38 for Validation, while the remaining 16 images are used for training. Table \ref{table:3} provides detailed information about each dataset.

\begin{table}[!ht]
  \begin{center}
  \scriptsize
    \caption{THE DETAILS OF DIFFERENT SEMANTIC SEGMENTATION DATASETS.}
    \begin{tabular}{c|c c c c c}
    \hline
      \text{Datasets} & \text{Training} & \text{Validation} & \text{Testing} & \text{Category} & \text{Input Size}\\
      \hline
      LoveDA & 2,522 & 1,669 & 1,796 & 7 &1,024$\times$1,024\\
      Potsdam & 24 & - & 14 & 6 &6,000$\times$6,000\\
      Vaihingen & 16 & - & 17 & 6 & 2,494$\times$2,064 \\
      \hline
    \end{tabular}
  \end{center}
  \label{table:3}
\end{table}

\subsection{Implementation Details and Evaluation Metrics}
\subsubsection{Implementation Details}
For the LoveDA dataset, the training and validation sets are both used for training. These images are cropped into patches with 512$\times$512 resolution for input. During training, various enhancement techniques such as random vertical flip, random horizontal flip, and random scaling with ratios of [0.5, 0.75, 1.0, 1.25, 1.5] are employed. The training process last for 400 epochs with a batch size of 16. During the testing phase, we use 1796 images provided by the official for prediction. As for the Potsdam and Vaihingen datasets, the images are cropped into 512$\times$512 for model input. The training epoch set to 200 with a batch size of 16.

In the experiment, learning rate warmup combined with cosine annealing is used to adjust the learning rate, where warmup set to 3 epochs. Moreover, AdamW~\cite{loshchilov2017decoupled} optimizer was selected to accelerate model convergence, with the learning rate and weight decay set to \num{1e-4} and \num{1e-8} respectively. All models are trained using NVIDIA GTX 3090 GPUs and implemented on the PyTorch framework.

\subsubsection{Evaluation Metrics}
We evaluate models based on two aspects: accuracy and performance. In terms of model accuracy, we employ metrics including the mean F1 score (F1), overall accuracy (OA), and mean intersection over union (mIoU). Regarding performance, we assess the model size using the number of model parameters (M), the model complexity through GPU memory usage (MB) and the floating point operation count (FLOPs), and the inference speed by measuring frames per second (FPS).

\subsection{Ablation Study and Comparison Experiments}

\subsubsection{Architecture Analysis}
To determine the foundational network architecture of Hi-ResNet, we conduct a series of ablation studies on the Vaihingen dataset. For fair comparison, all ablation studies share the same settings, except for the experimental variable. Table \ref{table:4-1} shows the experimental setup for architecture analysis.

\begin{table}[!ht]
  \renewcommand{\arraystretch}{1.8}
  \begin{center}
  \footnotesize
  \caption{EXPRIMENTAL SETUP FOR THE DIFFERENT ARCHITECTURE}
  \resizebox{\columnwidth}{!}{
  \begin{threeparttable}
    \begin{tabular}{c|c | c |c |c}
    \hline
        \multirow{2}{*}{Method} &  \multicolumn{4}{c}{Multi-branch Module}\\
        \cline{2-5}
          & Stage1 & Stage2 & Stage3 & Stage4\\
      \hline
       HRNet & $[Block]\times ^{\scriptscriptstyle1}B_{4}$ & $[Block]\times B_{4}\times ^{\scriptscriptstyle2}M_{1}$  & $[Block]\times B_{4}\times M_{4}$  & $[Block]\times B_{4}\times M_{3}$\\
       \hline
       V1 & $^{\scriptscriptstyle3}$  & &$[Block]\times \mathbf{B_{12}} \times M_{4}$\\
       \hline
       V2 &  &  & & $[Block]\times \mathbf{B_{12}} \times M_{3}$\\
       \hline
       Hi-ResNet base & & & $[Block]\times B_{12} \times M_{4}$ & \ding{56}\\
      \hline
    \end{tabular}
    \begin{tablenotes}
        \footnotesize  
        \item[1] 4 blocks are represented by $B_{4}$. 
        \item[2] 1 Module are represented by $M_{1}$.
        \item[3] Blank refers to keep the same settings of the baseline.
    \end{tablenotes}
    \end{threeparttable}
    }
  \end{center}
  \label{table:4-1}
\end{table}

Given that the overall structure of this paper inspires from HRNet~\cite{wang2020deep}, the baseline architecture is the same as the original HRNet, which consists of four stages. These four stages contain 1 to 4 branches respectively, each branch is composed of several high resolution modules with stacks of bottleneck blocks or basic blocks. Apart from the baseline, we set V1 to extends stage3 threefold, while V2 tripled stage4. Finally, our unique Hi-ResNet base not only triples the length of stage3 but also eliminates stage4 entirely.

Table \ref{table:4-2} displays the results of the ablative experiments. Surprisingly, although extending stage4 leads to more than doubling the number of network parameters, the mIoU increased by only 0.8\%, which is less than the mIoU obtained by extending stage3 with less increase in parameters. Through sampling analysis of the original stage3 and stage4 feature output images, we argue that there is some feature loss when extracting information in stage3, thereby reducing the medium and low-resolution semantic information that stage4 can acquire. Lengthening stage3 effectively addresses this issue, allowing for the extraction of richer and more accurate spatial information and better fitting of the features of HRS images in tasks. Consequently, we decide to lengthen stage3 by three times.
After determining the stage size, we attempt to eliminate redundant stage within the framework. We argues that lengthening stage3 can fully encompass the information extracted by stage4, so we try to remove stage4. This decision significantly reduces the number of parameters and FLOPs, speeds up the training process, and further enhances the model's efficiency and accuracy.
\begin{table}[!ht]
\renewcommand{\arraystretch}{1.2}
  \begin{center}
  \footnotesize
  \begin{threeparttable}
    \caption{RESULTS OF THE ARCHITECTURE ANALYSIS EXPERIMENTS ON VAIHINGEN DATASET}
    \begin{tabular}{c|c c c c c}
    \hline
      \text{Method} & \text{Params (M)} & \text{FLOPs (G)} & \text{Memory (MB)} & \text{mIoU}\\
      \hline
       HRNet & 68.6 & 140 & \textbf{1569} & 69.3\\
       V1 & 96.5& 205 & 2177 & 70.4\\
       V2 & 153.3 & 214 & 2379 & 70.1\\
       Hi-ResNet base & \textbf{46.5} & \textbf{139}  & 2116 & \textbf{70.8}\\
      \hline
    \end{tabular}
    \end{threeparttable}
  \end{center}
  \label{table:4-2}
\end{table}

Ultimately, we expanded the third stage of HRNet threefold and removed the fourth stage to establish the foundational architecture for Hi-ResNet. Different from the four-stage network architecture composed of HRNet with a fixed number of convolutions, our network extends specific stages to ensure it acquires more high-level semantic information. Compared to the baseline, Hi-ResNet base achieves more than 1.5\% increase in mIoU on the Vaihingen dataset and reduces the number of parameters by 30\%. In addition, in order to align the various modules proposed in Hi-ResNet, we renamed stage1 as the funnel stem, and stage2 and stage3 are referred to as layer1 and layer2, respectively.

\subsubsection{Module Analysis}
To evaluate the performance of each proposed component in Hi-ResNet, we cast ablation experiments on our modified modules such as Funnel Module, Multi-branch Module, Feature Refinement Module, and CEA loss, which bases on the Hi-ResNet base model. To prevent overfitting during the training process, all ablation studies employ both GD and LSCE loss, with a balanced weighting ratio of 1:1.

\begin{table}[!ht]
\caption{RESULTS OF THE MODULE ANALYSIS EXPERIMENTS ON VAIHINGEN DATASET}
\renewcommand{\arraystretch}{1.2}
  \begin{center}
  \footnotesize
  \resizebox{\columnwidth}{!}{
  \begin{threeparttable}
    \begin{tabular}{c|c c c c c c}
    \hline
      \text{Method} & \text{Funnel} & \text{Multi-branch} & \text{Feature Refinement} & \text{CEA} & \text{Params} & \text{mIoU}\\
      \hline
       Hi-ResNet base & &  & &  & 46.5& 71.0 \\
       Hi-ResNet v1 &\ding{52}  &  &  & & 50.2& 72.2\\
       Hi-ResNet v2 & \ding{52}  &  \ding{52} &   &  & \textbf{22.3}&74.7\\
       Hi-ResNet v3 & \ding{52}  &\ding{52}   & \ding{52}  &  & 26.0& 75.2\\
       ours & \ding{52}  & \ding{52}  & \ding{52}  & \ding{52}  & 26.0 &\textbf{76.2}\\
      \hline
    \end{tabular}
    \end{threeparttable}
    }
  \end{center}
  \label{table:5}
\end{table}

As shown in Table \ref{table:5}, Hi-ResNet base acquires only 71.0\% of mIoU, indicating a limited ability to segment HRS images. The addition of the funnel module increases model precision by 1.2\%, demonstrating its effectiveness in capturing more high-resolution semantic information during the downsampling process. Notably, the muti-branch module greatly reduces the number of parameters of the model (more than 50 percent), while providing a significant increase of at least 2.5 \% for the model. This fully demonstrates the excellent performance of IA block, which can effectively suppress the interference of irrelevant background and extract more accurate feature information. This also illustrates the effectiveness of feature fusion. By feature refinement module, the model outputs both coarse and refined segmentations. However, the model's mIoU only increases by 0.5\%. This is because the combination of GD and LSCE loss has difficulty in utilizing the two outputs of the model. Therefore, we add the CEA loss to the model training to form the final Hi-ResNet. Without any increase in additional parameters, CEA loss improves model precision by 1\%, proving its capacity to balance model outputs effectively and its compatibility with the model. Compared to Hi-ResNet base, the final Hi-ResNet achieves 76.2\% mIoU without pre-training, demonstrating the effectiveness of each components of the model.

\subsubsection{Loss Analysis}
We use LSCE, GD, and CEA loss for training the Hi-ResNet, aiming to improve the accuracy and generalization ability of the model. Nevertheless, imbalanced weights among the various loss functions could potentially instigate gradient conflicts and destabilize the training process. Therefore, to eliminate this instability, we conduct ablation studies on the weights between the loss functions across the Potsdam, Vaihingen and LoveDA datasets. All loss calculations in ablation studies were performed exclusively with the results from the refined segmentation.

We sum the GD, LSCE, and CEA loss with a default 1:1:1 ratio as V1. In the V2, we weight the LSCE, GD, and CEA loss in a 1:1:0.4 ratio to ensure that the values of the three loss functions are in the same scale. Subsequently, we select Random Weighting~\cite{lin2021reasonable} as V3. This approach employs dynamic loss weights in each training iteration, samples loss weights from a potentially normalized distribution, and minimizes the aggregate loss weighted by these randomly sampled weights. Finaly, we take the softmax computation result of 1:1:0.4, i.e., (0.3923, 0.3923, 0.2153) as V4.
\begin{table}[!ht]
  \begin{center}
  \caption{RESULTS OF LOSS ANALYSIS ON THE THREE DATASETS}
  \renewcommand{\arraystretch}{1.1}
  \footnotesize
    \setlength{\tabcolsep}{6mm}{
    \begin{threeparttable}
    \begin{tabular}{c|c c c c c}
      \hline
         Method & LoveDA & Potsdam & Vaihingen\\
        \hline
      V1 & $^{\scriptscriptstyle1}$49.7 & 81.8 & 75.1\\
      V2 & 50.0 & 82.3 & 75.7\\
      V3 & 49.8 & 82.0 & 75.3\\
      V4 (Ours) & \textbf{50.1} & \textbf{82.5} & \textbf{75.8}\\
      \hline
    \end{tabular}
    \begin{tablenotes}
        \footnotesize  
        \item[1] All number in this rectangle refers to mIoU. 
      \end{tablenotes} 
    \end{threeparttable}
    }
  \end{center}
  \label{table:6}
\end{table}
    
As shown in Table \ref{table:6}, V1 achieves the lowest mIoU on all three datasets, indicating that the unreasonable weighting of the loss functions hinders model optimization. It is noteworthy that the mIoU obtained with fixed weights V2 and V4 are both higher than the mIoU obtained with the Random Weighting strategy. This is because although the use of dynamically weighted loss gives the model more diversity, it ignore the optimal weight of the loss. Meanwhile, compared to V2, V4 after softmax function performes best on all three datasets. This suggests that normalized loss weights can effectively utilize all loss functions, avoiding instability in training due to improper weight settings. We apply the optimal weights of V4 to the model.

\begin{figure}[h!]
    \centering
    \includegraphics[width=\linewidth]{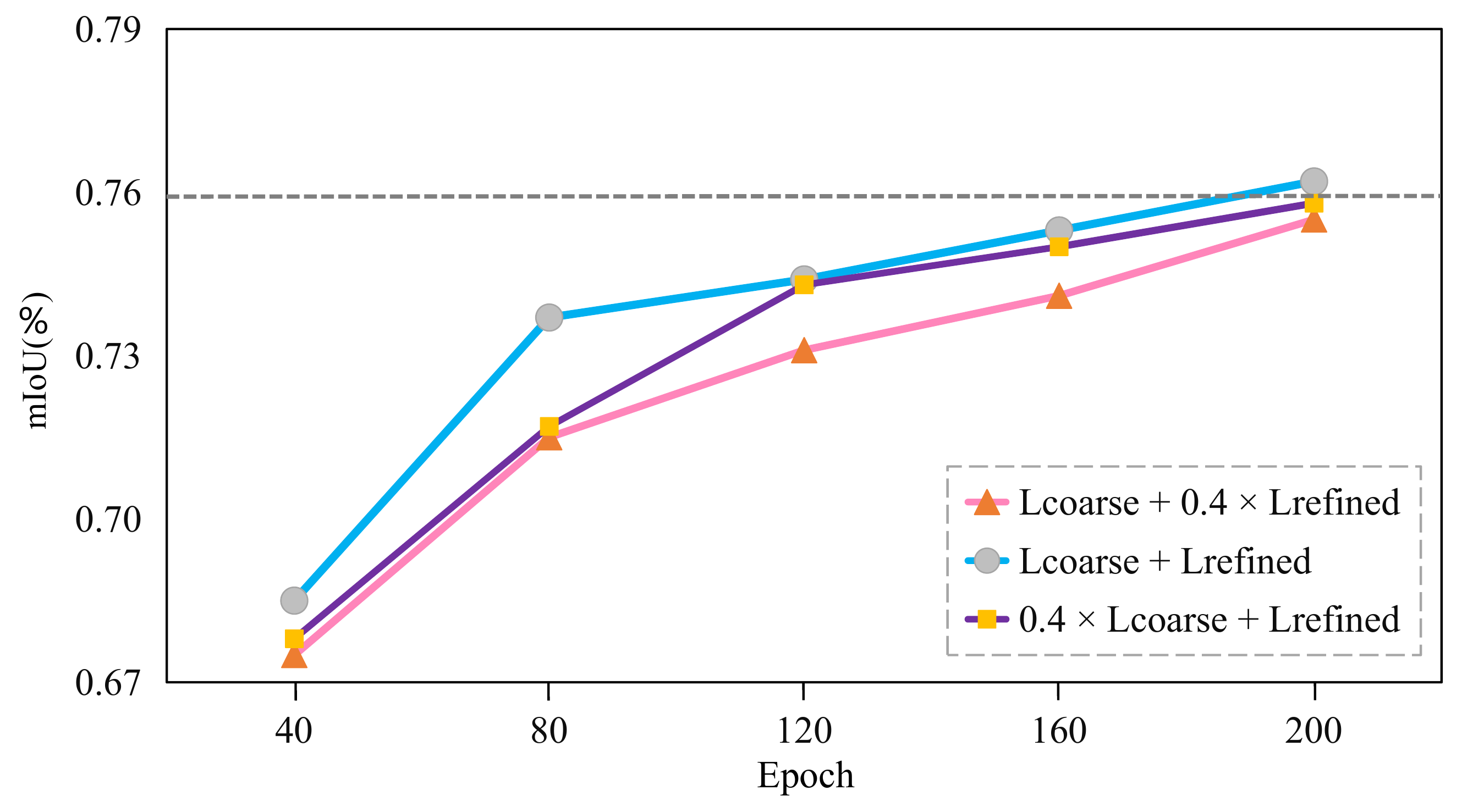}
    \caption{Results on the Vaihingen dataset using different ratios to weight the losses of coarse segmentation and refined segmentation.}
    \label{fig:loss_ratio}
\end{figure}

As mentioned before, Hi-ResNet finally outputs two results: coarse segmentation and refined segmentation. These two outputs will calculate three losses respectively according to the conclusion in Table \ref{table:6}. However, what weight to add up the losses of the two outputs is still a question we need to consider. Therefore, we test different weights for the losses of the two outputs. Specifically, we add the loss of coarse segmentation and the loss of refined segmentation using weights of 1:0.4, 1:1, and 0.4:1 respectively (note that the ratio of 1:0.4 is the same as the original OCR setup~\cite{yuan2019segmentation}). As shown in Figure \ref{fig:loss_ratio}, it can be seen that when the losses of the two outputs are added in a 1:1 ratio, the model can achieve the highest mIoU of 76.2\%, which is superior to the settings of the original OCR network. This suggests that the target location information contained in the coarse segmentation and the object edge information contained in the refined segmentation can complement each other well, guiding the optimization direction of the model together, thereby improving the model's accuracy.

\subsubsection{Pre-training Comparison}
To illustrate the impact of pre-training strategies on downstream HRS tasks, we separately finetune the two pretrained models on three datasets. For supervised pre-training, the original Mapillary dataset~\cite{neuhold2017mapillary} is randomly cropped into 256$\times$256 pixels, and 400,000 category-balanced HRS images are selected as the pre-training dataset. The batch size is set to 80, and the base learning rate is set to \num{5e-5}. The maximum iteration number of this pre-training is 3,125,000, achieving 51.8 on the Mapillary validation set. For unsupervised pre-training, the MillionAID dataset~\cite{long2021creating} with randomly cropped 400$\times$400 pixels is used. The learning rate is set to 0.015, the batch size is 64, and the maximum iteration number is the same as supervised pre-training. The final top-1 accuracy on MillionAID is 78.9, and the top-5 accuracy is 94.1. We conduct a comprehensive evaluation of HRS pre-training models on three datasets, and the detailed information presentes in Table \ref{table:7}.

\begin{table}[!ht]
  \begin{center}
  \renewcommand{\arraystretch}{1.1}
  \begin{threeparttable}
   \setlength{\tabcolsep}{3mm}{
    \caption{RESULTS OF THE PRETRAINING COMPARISON EXPERIMENTS}
    \begin{tabular}{c|c | c c c c c}
      \hline
        Method & Max iteration & LoveDA & Potsdam & Vaihingen\\
        \hline
      $^{\scriptscriptstyle1}$NP & - & $^{\scriptscriptstyle2}$50.4 & 82.8 & 76.2\\
      SP & 3,125,000 & 51.8 & 85.2 & 78.6\\
      MoCo & 3,125,000 & \textbf{52.6} & \textbf{87.6} & \textbf{79.8}\\
      \hline
    \end{tabular}}
    \begin{tablenotes}
        \footnotesize  
        \item[1] NP: No pre-training, SP: supervised pre-training, MoCo: unsupervised pre-training using MoCoV2. 
        \item[2] All number in this rectangle refers to mIoU.
        
      \end{tablenotes} 
    \end{threeparttable}
  \end{center}
  \label{table:7}
\end{table}

Two pre-trained model weights are loaded onto the Hi-ResNet, and the result shows that unsupervised pre-training of HRS images using MoCoV2~\cite{he2020momentum}  can provides a 3.6\% increase in mIoU, while supervised pre-training only increases mIoU by 2.4\% under the same number of iterations. Therefore, we use the unsupervised pre-training weights in the subsequent experiments. The experiment demonstrates that unsupervised remote sensing pre-training can significantly improve the performance of the model on a small data set and make the model converge faster. Furthermore, using MoCoV2 provides a deeper feature representation for HRS downstream tasks. In this sense, pre-trained models using contrastive learning methods can offer competitive backbones for future research in the field of HRS.

Moreover, due to limited computing resources, we perform only 3-4 complete pre-training processes (150 epochs), which makes pre-training the depth provided by Hi-ResNet compares with the pre-training weights provided by other officials, there is a certain gap between the hierarchical representation information. However, the result on the LoveDA dataset in Table \ref{table:10} fully demonstrates the excellent performance of Hi-ResNet in terms of performance and accuracy.

\subsubsection{Model Efficiency Comparison}
In addition to accuracy and precision, the complexity and speed of a model are equally important for HRS tasks. Therefore, we use a single NVIDIA GTX 3090 GPU to compare our proposed model with classic CNN-based networks and Transformer-based networks in terms of the model parameters, GPU memory usage, and FLOPs. We chose to train on the large-scale, i.e., 1024$\times$1024 LoveDA dataset, and the comparison results can be seen in Table \ref{table:8}.

\begin{table}[!ht]
\renewcommand{\arraystretch}{1.1}
  \scriptsize
  \begin{center}
   \setlength{\tabcolsep}{1mm}{
    \caption{EFFICIENCY OF THE DIFFERENT NETWORKS ON THE LOVEDA DATASET}
    \begin{tabular}{c|c c c c c}
    \hline
      \text{Method} & \text{backbone} & \text{Params(M)} & \text{Memory(MB)} & \text{FLOPs(G)} & \text{mIoU}\\
      \hline
      Segmenter~\cite{strudel2021segmenter} & ViT-T & \textbf{6.7} & 3495 & \textbf{26} & 47.1\\
      UperNet~\cite{wang2022advancing} & ViT-B + RVSA & 114.0 & 25343 & 407 & 51.9\\
      SegFormer~\cite{xie2021segformer} & MiT-B1 & 13.7 & 3933 & 63 & 51.1\\
      DeepLabV3+~\cite{chen2018encoder} & ResNet50 & 59.3 & \textbf{1063} & 355 & 47.6 \\
      HRNet~\cite{wang2020deep} & HRNet-W48 & 75.9 & 1969 & 559 & 49.8 \\
      \hline
      Hi-ResNet & Hi-ResNet & 26.0 & 2116 & 402 & \textbf{52.6} \\
      \hline
    \end{tabular}}*-
  \end{center}
  \label{table:8}
\end{table}

Compared with the classic CNN based DeeplabV3+~\cite{chen2018encoder}, our model has relatively higher memory usage and FLOPs. This is because Hi-ResNet maintains the high resolution of the input, while simultaneously extracting image features at multiple resolutions in parallel. However, our model has almost half the number of parameters compared to DeeplabV3+, while achieving a nearly 5\% higher mIoU. It is worth mentioning that although HRNet~\cite{wang2020deep} uses the same parallel structure as our proposed model, this structural improvement in our ablation studies allows our model to achieve higher mIoU with fewer parameters as well as FLOPs. Due to the global attention mechanism, Transformer-based semantic segmentation networks like SegFormer~\cite{xie2021segformer} often require expensive computational resources, while having a relatively smaller number of parameters. In contrast, our model maintains a balance between GPU memory and FLOPs, allowing it to achieve superior accuracy within a reasonable complexity.

\begin{table}[!ht]
  \begin{center}
  \begin{threeparttable}
   \setlength{\tabcolsep}{2mm}{
    \caption{RESULTS OF DIFFERENT INPUT SIZE ON THE VAIHINGEN DATASET}
    \begin{tabular}{c|c c c c c c c}
    \hline
     \text{Input Size} & $^{\scriptscriptstyle1}$\text{Imp.surf} & \text{Building} & \text{Lowveg} & \text{Tree} & \text{Car} & \text{mIoU}\\
      \hline
      256x256 & $^{\scriptscriptstyle2}$85.3 & 90.1 & 72.1 & 80.2 & 70.1 & 79.5 \\
      256x512 & 84.1 & 90.1 & 71.1 & 80.2 & 69.9 & 79.1\\
      512x512 & 85.5 & 90.6 & 72.5 & 80.3 & 70.2 & \textbf{79.8}\\
      512x1024 & 84.8 & 90.2 & 71.8 & 80.1 & 69.8 & 79.3\\
      1024x1024 & 85.5 & 90.2 & 72.1 & 80.1 & 69.9 & 79.6\\
      \hline
    \end{tabular}}
        \begin{tablenotes} 
        \footnotesize  
        \item[1] Imp. surf: impervious surfaces, Lowveg: low vegetation.
        \item[2] All number in this rectangle refers to mIoU.
        
      \end{tablenotes} 
      \end{threeparttable}
  \end{center}
  \label{table:9}
\end{table}

\subsubsection{Stability Analysis} 
To validate the stability of the proposed model, we conduct experiments on the Vaihingen dataset using various input sizes, including square sizes of 256$\times$256, 512$\times$512, and 1024$\times$1024, as well as rectangular sizes of 256$\times$512 and 512$\times$1024.

\begin{table*}[!t]
  \scriptsize
  \begin{center}
  \begin{threeparttable}
   \setlength{\tabcolsep}{2.2mm}{
    \caption{PERFORMANCE OF THE REFERENCE METHODS AND THE PROPOSED 
Hi-ResNet METHOD ON THE LOVEDA DATASET}
    \begin{tabular}{c|c c c c c c c c c c c c}
    \hline
        \multirow{2}{*}{Method} & \multirow{2}{*}{Backbone} & \multirow{2}{*}{Pretrain} & \multicolumn{8}{c}{IoU per class(\%)}  \multirow{2}{*}{mIoU} & \multirow{2}{*}{FLOPs(G)} & \multirow{2}{*}{FPS} \\
        \cline{4-10} 
       &&& \text{Background} & \text{Building} & \text{Road} & \text{Water} & \text{Barren} & \text{Forest} & \text{Agriculture} \\
      \hline
       PSPNet~\cite{zhao2017pyramid} & ResNet50& Y & 44.4 & 52.1 & 53.5 & 76.5 & 9.7 & 44.1 & 57.9 & 48.3 & 738 & 27\\
       DeepLabV3+~\cite{chen2018encoder}& ResNet50& Y & 43.0 & 50.9 & 52.0 & 74.4 & 10.4 & 44.2 & 58.5 & 47.6 & 355 & 46 \\
       UNet++~\cite{zhou2018unet++} & ResNet50& Y & 42.8 & 52.6 & 52.8 & 74.5 & 11.4 & 44.4 & 58.8 & 48.1 & 544 & 30\\
       SemanticFPN~\cite{kirillov2019panoptic} & ResNet50& Y & 42.9 & 51.5 & 53.4 & 74.7 & 11.2 & 44.6 & 58.7 & 48.2 & 589 & 37 \\
       FarSeg~\cite{zheng2020foreground}& ResNet50& Y & 43.1 & 51.5 & 53.9 & 76.6 & 9.8 & 43.3 & 58.9 & 48.2 & 350  & 47\\
       BANet~\cite{wang2021transformer} & ResT-Lite& Y & 43.7 & 51.5 & 51.1 & 76.9 & 16.6 & 44.9 & 62.5 & 49.6 & 67  &  \textbf{84}\\
       TransUNet~\cite{chen2021transunet} & ViT-R50& Y & 43.0 & 56.1 & 53.7 & 78.0 & 9.3 & 44.9 & 56.9 & 48.9 & 803  & 13\\
       Segmenter~\cite{strudel2021segmenter}& ViT-Tiny& Y & 38.0 & 50.7 & 48.7 & 77.4 & 13.3 & 43.5 & 58.2 & 47.1 & \textbf{26}  & 61\\
       SwinUperNet~\cite{liu2021swin} & Swin-Tiny& Y & 43.3 & 54.3 & 54.3 & 78.7 & 14.9 & 45.3 & 59.6& 50.0 & 349  & 19\\
       FactSeg~\cite{ma2021factseg} & ResNet50& Y & 42.6 & 53.6 & 52.8 & 76.9 & 16.2 & 42.9 & 57.5 & 48.9 & 267  & 46\\
       DC-Swin~\cite{wang2022novel} & Swin-Tiny& Y & 41.3 & 54.5 & 56.2 & 78.1 & 14.5 & 47.2 & 62.4 & 50.6 & 107  & 60\\
       UperNet~\cite{wang2022advancing}& VITAE-B + RVSA& Y & 46.7 & 58.1 & 57.1 & 79.6 & 16.5 & 46.4 & 62.4 & 52.4 & 413 & 11\\
       UperNet~\cite{wang2022advancing}& VIT-B + RVSA& Y & 45.2 & 59.8 & 55.2 & 79.4 & 18.4 & 46.2 & 59.2 & 51.9 & 407 & 19\\
       RSSFormer~\cite{xu2023rssformer}& RSS-B& Y & 52.3 & 60.7 & 55.2 & 76.2 & 18.7 & 45.3&  58.3 & 52.3 & 413  & 6\\

       AerialFormer-S~\cite{yamazaki2023aerialformer}& AerialFormer & Y & 46.6 & 57.4 & 57.3 & 80.5 & 15.6 & 46.8 & 62.8 & 52.4 & - & - \\
      \hdashline
      HRNet(Baseline)~\cite{wang2020deep} & HRNet-W48 & Y & 44.6 & 55.3 & 57.4 & 78.0 & 11.0 & 45.3 & 60.9 & 49.8 & 559 & 25\\
      Hi-ResNet & Hi-ResNet & Y & 46.8 & 58.3 & 55.9 & 80.1 & 17.0 & 46.7 & 62.7 & \textbf{52.6} & 402 & 35\\
      \hline
    \end{tabular}}
    \end{threeparttable}
  \end{center}
  \label{table:10}
\end{table*}

\begin{figure*}[t!]
    \centering
    \includegraphics[width=\linewidth]{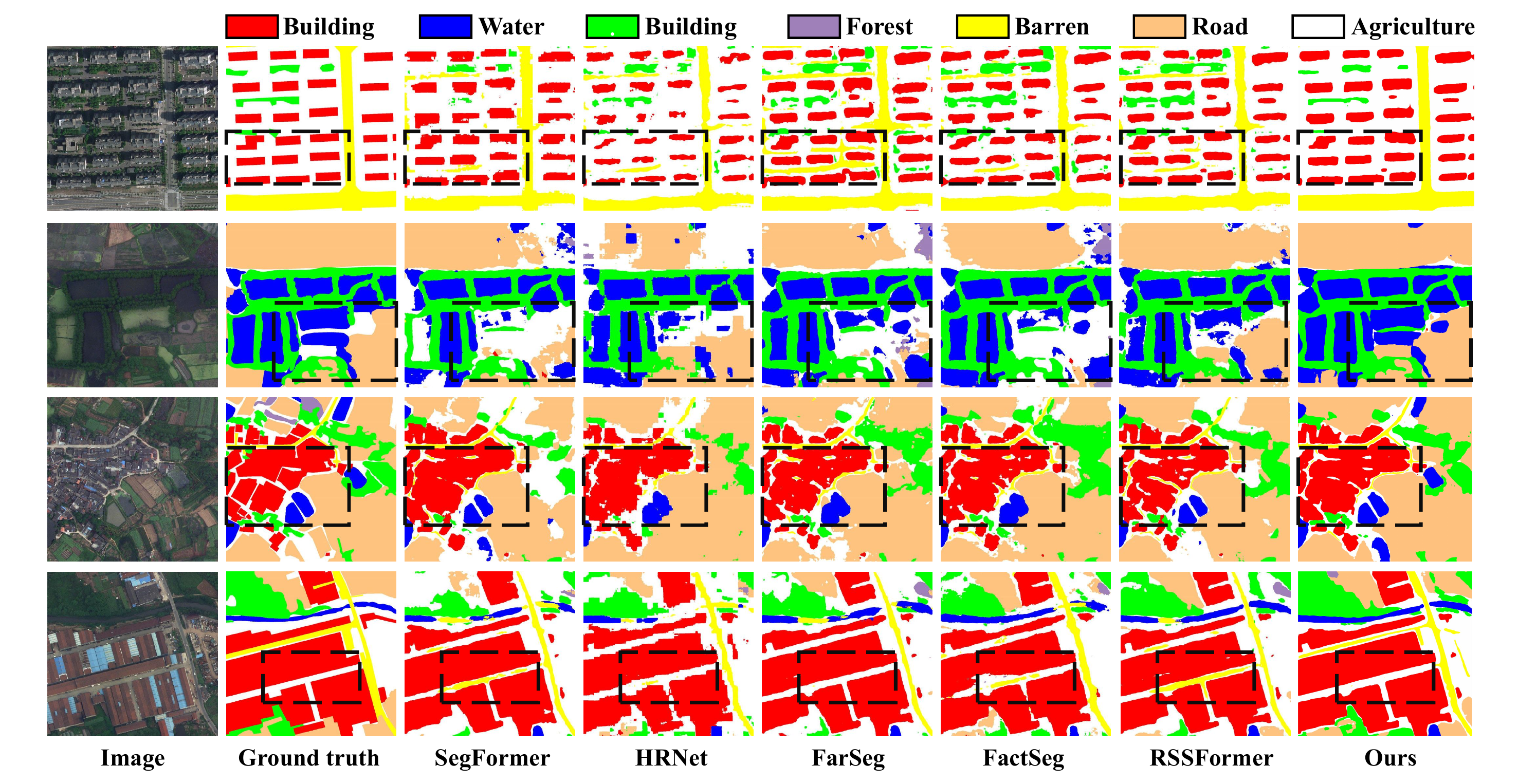}
    \caption{Visual results of different methods on the LoveDA dataset. From left to right: original image, ground truth, results of SegFormer~\cite{xie2021segformer}, results of HRNet~\cite{wang2020deep}, results of FarSeg~\cite{zheng2020foreground}, results of FactSeg~\cite{ma2021factseg}, results of RSSFormer~\cite{xu2023rssformer}, and results of our Hi-ResNet.}
    \label{fig:loveda}
\end{figure*}

\begin{table*}[!t]
  \scriptsize
  \begin{center}
  \begin{threeparttable}
   \setlength{\tabcolsep}{2.8mm}{
    \caption{PERFORMANCE OF THE REFERENCE METHODS AND THE PROPOSED
Hi-ResNet METHOD ON THE POTSDAM DATASET}
    \begin{tabular}{c|c c c c c c c c c c c c}
    \hline
      \multirow{2}{*}{Method} & \multirow{2}{*}{Backbone} & \multirow{2}{*}{Pretrain} & \multicolumn{6}{c}{F1 per class(\%)} \multirow{2}{*}{MeanF1} & \multirow{2}{*}{OA} & \multirow{2}{*}{mIoU} & \multirow{2}{*}{FLOPs(G)} & \multirow{2}{*}{FPS}\\
        \cline{4-8} 
       &&& $^{\scriptscriptstyle1}$\text{Imp.surf} & \text{Building} & \text{Lowveg} & \text{Tree} & \text{Car}\\
       
      \hline
      ERFNet~\cite{romera2017erfnet} & ERF & Y & 88.7 & 93.0 & 81.1 & 75.8 & 90.5 & 85.8 & 84.5 & 76.2 & \textbf{11} & 142\\`  
      BiSeNet~\cite{yu2018bisenet} & ResNet18 & Y & 90.2 & 94.6 & 85.5 & 86.2 & 92.7 & 89.8 & 88.2 & 81.7 & 20 & 130\\
      DANet~\cite{fu2019dual} & ResNet18 & Y & 89.9 & 93.2 & 83.6 & 82.3 & 92.6 & 88.3 & 86.7 & 79.6 & 58 & \textbf{157}\\
      ShelfNet~\cite{zhuang2019shelfnet} &  ResNet18 & Y & 92.5 & 95.8 & 86.6 & 87.1 & 94.6 & 91.3 & 89.9 & 84.4 & 98 & 123\\
      FANet~\cite{hu2020real} & ResNet18 & Y & 92.0 & 96.1 & 86.0 & 87.8 & 94.5 & 91.3 & 89.8 & 84.2 & 79 & 118\\
       Segmenter~\cite{strudel2021segmenter}& ViT-Tiny & Y & 91.5 & 95.3 & 85.4 & 85.0 & 88.5 & 89.2 & 88.7 & 80.7 & 12 & 138\\
       SwinUperNet~\cite{liu2021swin} & Swin-Tiny & Y & 93.2 & 96.4 & 87.6 & 88.6 & 95.4 & 92.2 & 90.9 & 85.8 & - & -\\

       UperNet~\cite{wang2022advancing}& ResNet50 & Y & 92.4 & 96.1 & 85.7 & 85.5 & 89.9 & 89.9 & 90.6 & - & - & -\\
       UperNet~\cite{wang2022advancing}& Swin-Tiny & Y & 92.6 & 96.3 & 86.0 & 85.4 & 89.7 & 90.1 & 90.8 & - & 215 & 58\\
       DC-Swin~\cite{wang2022novel}& Swin-Tiny & Y & 94.2 & 97.6 & 88.6 & 89.6 & 96.3 & \textbf{93.3} & 92.0 & 87.5 & 23 & 72\\
       BSNet~\cite{hou2022bsnet}& ResNet50 & Y & 92.4 & 95.6 & 86.8 & 88.1 & 94.6 & 91.5 & 90.7 & 77.5 & - & -\\
       ST-UNet~\cite{he2022swin}& ResNet50 & Y & - & - & - & - & - & 86.1 & - & 75.9 & - & 9\\
       RSSFormer~\cite{xu2023rssformer}& RSS-B & Y & 93.8 & 96.0 & 86.8 & 86.7 & 96.8 & 92.0 & 91.0 & - & 84 & 11\\
       EfficientUNets~\cite{almarzouqi2023semantic}& EfficientB7 & N & 91.5 & 96.3 & 79.4 & 90.9 & 88.1 & 89.2 & 90.8 & 80.5 & - & -\\
       UperNet~\cite{cha2023billion}& VIT-G & Y & 92.7 & 96.9 & 85.8 & 89.0 & 96.0 & 92.1 & 92.5 & - & - & -\\
        
      \hdashline
      HRNet(Baseline)~\cite{wang2020deep}& HRNet-W48 & Y & 88.7 & 93.4 & 83.0 & 81.5 & 91.1 & 87.5 & 86.1 & 78.1 & 279 & 121\\
      Hi-ResNet & Hi-ResNet & Y & 93.2 & 96.5 & 87.9 & 88.6 & 96.1 & 92.4 & \textbf{92.6} & \textbf{87.6} & 57 & 131\\
      \hline
    \end{tabular}}
    \begin{tablenotes}
        \footnotesize  
        \item[1] Imp. surf: impervious surfaces. Lowveg: low vegetation. 
      \end{tablenotes} 
    \end{threeparttable}
  \end{center}
  \label{table:11}
\end{table*}

\begin{figure*}[t!]
    \centering
    \includegraphics[width=\linewidth]{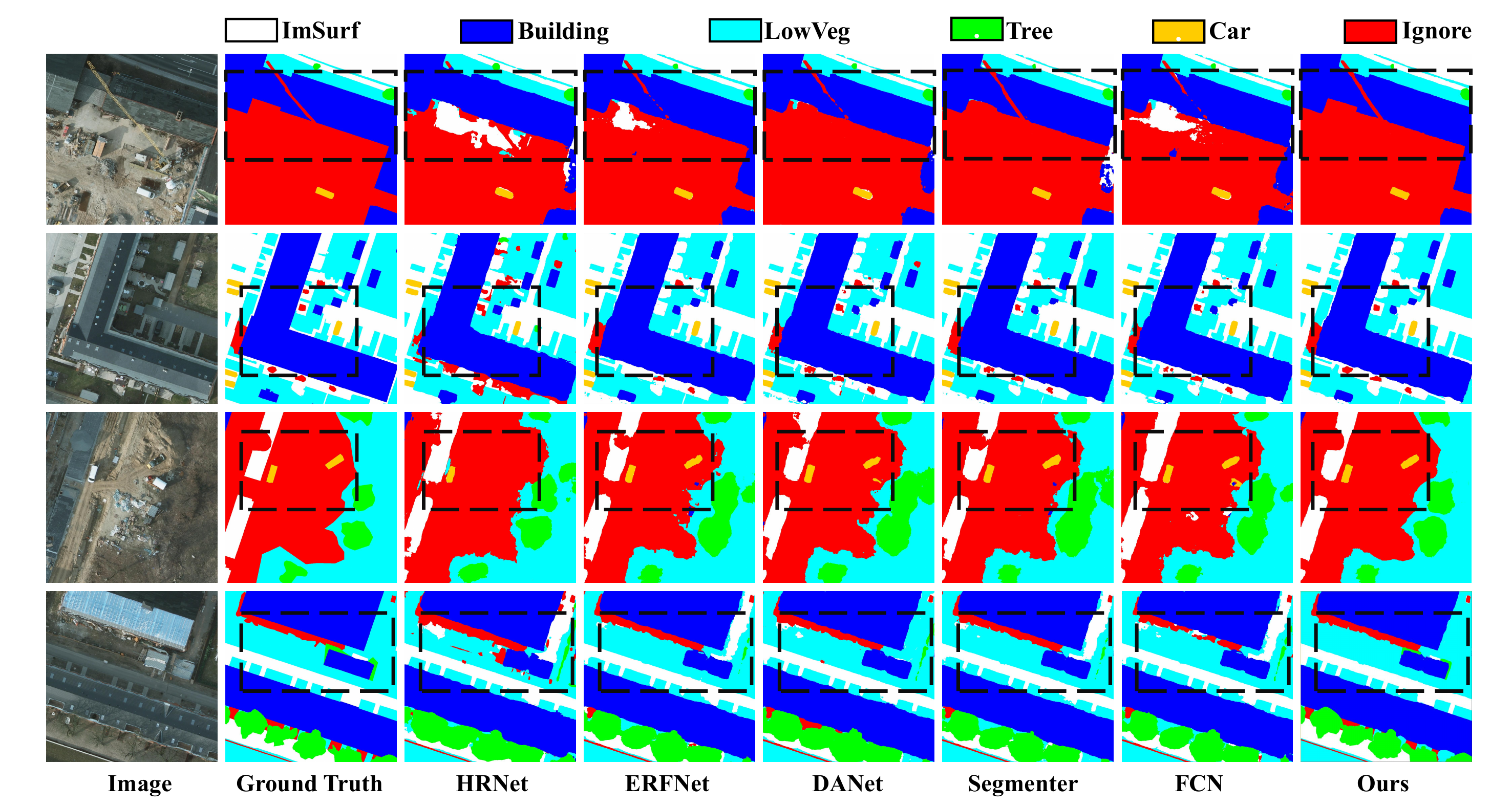}
    \caption{Visualization results for the Potsdam validation set. From left to right: original image, ground truth, results of HRNet~\cite{wang2020deep}, results of ERFNet~\cite{romera2017erfnet}, results of DANet~\cite{fu2019dual}, results of Segmenter~\cite{strudel2021segmenter}, results of FCN~\cite{long2015fully}, and results of our Hi-ResNet.}
    \label{fig:potsdam}
\end{figure*}

\begin{table*}[!t]
  \scriptsize
  \begin{center}
   \setlength{\tabcolsep}{2.8mm}{
    \caption{PERFORMANCE OF THE REFERENCE METHODS AND THE PROPOSED
Hi-ResNet METHOD ON THE VAIHINGEN DATASET}
    \begin{tabular}{c|c c c c c c c c c c c c}
    \hline
        \multirow{2}{*}{Method} & \multirow{2}{*}{Backbone} & \multirow{2}{*}{Pretrain} & \multicolumn{6}{c}{F1 per class(\%)} \multirow{2}{*}{MeanF1} & \multirow{2}{*}{OA} & \multirow{2}{*}{mIoU} & \multirow{2}{*}{FLOPs(G)} & \multirow{2}{*}{FPS}\\
        \cline{4-8} 
       &&& \text{Imp.surf} & \text{Building} & \text{Lowveg} & \text{Tree} & \text{Car}\\
      \hline
       PSPNet~\cite{zhao2017pyramid}& ResNet18 & Y & 89.0 & 93.2 & 81.5 & 87.7 & 43.9 & 79.0 & 87.7 & 68.6& 53 & 112 \\
       BiSeNet~\cite{yu2018bisenet} & ResNet18 & Y & 89.1 & 91.3 & 80.9 & 86.9 & 73.1 & 84.3 & 87.1 & 75.8 & 20 & 128\\
      DABNet~\cite{li2019dabnet}& DAB & N & 87.8 & 88.8 & 74.3 & 84.9 & 60.2 & 79.2 & 84.3 & 70.2& \textbf{7} & 146 \\
      DANet~\cite{fu2019dual}& ResNet18 & Y & 90.0 & 93.9 & 82.2 & 87.3 & 44.5 & 79.6 & 88.2 & 69.4& 58 & 153 \\
      ShelfNet~\cite{zhuang2019shelfnet}& ResT-Lite & Y & 91.8 & 94.6 & 83.8 & 89.3 & 77.9 & 87.5 & 89.8 & 78.3& 98 & 122 \\
      FANet~\cite{hu2020real}& ResNet18 & Y & 90.7 & 93.8 & 82.6 & 88.6 & 71.6 & 85.4 & 88.9 & 75.6& 79 & 118 \\
      EaNet~\cite{zheng2020parsing}& ResNet18 & Y & 91.7 & 94.5 & 83.1 & 89.2 & 80.0 & 87.7 & 89.7 & 78.7 & - & - \\
       ABCNet~\cite{li2021abcnet}& ResNet18 & Y & 92.7 & 95.2 & 84.5 & 89.7 & 85.3 & 89.5 & 90.7 & 81.3& 16 & \textbf{185} \\
       BoTNet~\cite{srinivas2021bottleneck}& BoTNet50 & Y & 89.9 & 92.1 & 81.8 & 88.7 & 71.3 & 84.8 & 88.0 & 74.3& 102 & - \\
       Segmenter~\cite{strudel2021segmenter}& ViT-Tiny & Y & 89.8 & 93.0 & 81.2 & 88.9 & 67.6 & 84.1 & 88.1 & 73.6& 12 & 130 \\
       
      BSNet~\cite{hou2022bsnet}& ResNet50 & Y & 92.1 & 94.4 & 83.1 & 88.3 & 86.7 & 88.9 & 90.3 & 80.2 & - & - \\ 
      DC-Swin~\cite{wang2022novel}& Swin-S & Y & 93.6 & 96.1 & 85.7 & 90.3 & 87.6 & \textbf{90.6} & 91.6 & \textbf{83.2} & 23 & 80 \\
      ST-UNET~\cite{he2022swin}& ResNet50 & Y & - & - & - & - & - & 82.1 & - & 70.2 & - & 7 \\
      RSSFormer~\cite{xu2023rssformer}& RSS-B & Y & 93.7 & 96.8 & 81.3 & 91.7 & 89.2 & 90.5 & 90.8 & - & 84 & 10 \\
      EfficientUNets~\cite{almarzouqi2023semantic}& EfficientB7 & N & 91.4 & 96.3 & 79.4 & 90.8 & 88.1 & 89.0 & 90.8 & 73.1 & - & - \\
      
      \hdashline
      HRNet(Baseline)~\cite{wang2020deep}& HRNet-W48 & Y & 89.8 & 92.8 & 81.0 & 86.8 & 79.5 & 86.0 & 87.6 & 75.8& 279 & 120 \\
      Hi-ResNet & Hi-ResNet & Y & 92.3 & 95.1 & 84.9 & 89.5 & 83.5 & 90.1 & \textbf{91.7} & 79.8& 57 & 123 \\
      \hline
    \end{tabular}}
  \end{center}
  \label{table:12}
\end{table*}

\begin{figure*}[t!]
    \centering
    \includegraphics[width=\linewidth]{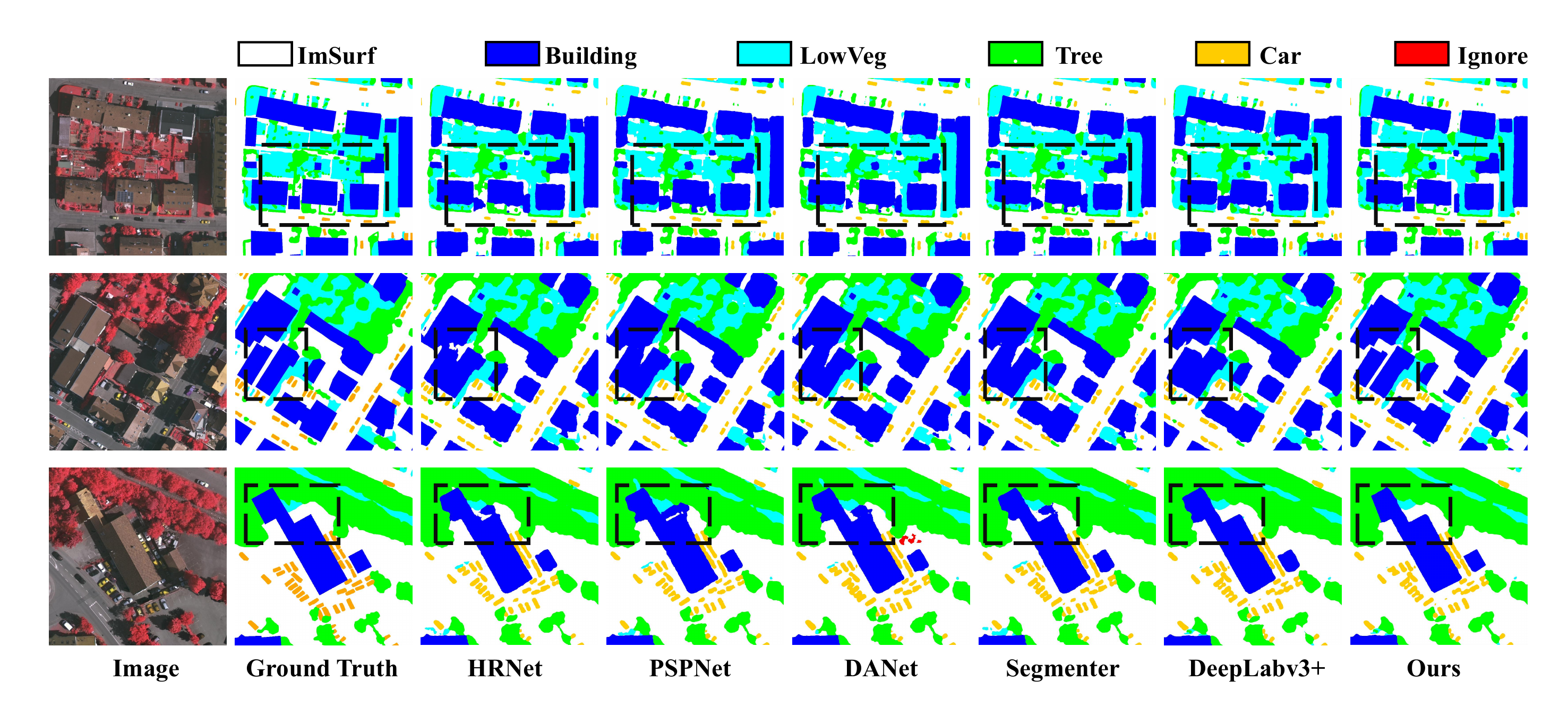}
    \caption{Visualization results for the Vaihingen validation set. From left to right: original image, ground truth, results of HRNet~\cite{wang2020deep}, results of PSPNet~\cite{zhao2017pyramid}, results of DANet~\cite{li2019dabnet}, results of Segmenter~\cite{strudel2021segmenter}, results of DeepLabv3+~\cite{chen2018encoder}, and results of our Hi-ResNet.}
    \label{fig:vaihingen}
\end{figure*}

As shown in Table \ref{table:9}, the Hi-ResNet presents a mIoU deviation of less than 0.8\% for inputs of different sizes, with the best performance observed for input size of 512$\times$512. When training with large-scale HRS images of 1024$\times$1024, our network maintains its accuracy on the ``Cars'' class, which substantiates its effectiveness in segmenting smaller objects within HRS images. Correspondingly, with an input of smaller 256$\times$256 images, the mIoU attained by the model is insignificantly different from the optimal results, suggesting that the proposed model has a larger receptive field. Furthermore, even with rectangular inputs such as 512$\times$1024, the model still attaines the mIoU exceeding 79\%, demonstrateing the stability of Hi-ResNet.

\subsection{Results on The Dataset}

\subsubsection{LoveDA}
The LoveDA dataset is recognized as a challenging HRS dataset for land cover domain adaptive semantic segmentation. This dataset presents three significant challenges for large-scale remote sensing mapping, namely multi-scale targets, complex background samples, and inconsistent class distributions. As a result, achieving high scores on this dataset is quite difficult.

Table \ref{table:10} demonstrates the results of different methods on the LoveDA dataset, where both FPS and FLOPs are evaluated on a single NVIDIA GTX 3090 GPU with an input size of 1024$\times$1024. In addition, the official backbone pre-training weights are used for all the networks in Table \ref{table:10}. Thanks to the precise loss strategy, we can handle complex samples of different backgrounds well on LoveDA and achieve the mIoU of 52.5\% on the official test set. Our network outperforms the HRNet~\cite{wang2020deep}, loaded with officially provided pre-trained weights, by 2.5\% on mIoU and FactSegNet~\cite{ma2021factseg}, an excellent small-object semantic segmentation network, by 3.5\%. It is worth noting that for the ``Barren'' class, where most networks underperform, our mIoU is 3\% higher than most methods. Whether in urban or rural scenarios, sparse or dense distribution, our network can accurately segment objects with high confidence.

However, We admit that compare to models like Segmenter~\cite{strudel2021segmenter}, Hi-ResNet underperforms in terms of model complexity and FPS. We attribute it to two reasons. First, the use of high-resolution 1024$\times$1024 input images significantly increases the computation of our model's attention mechanism, which in turn reduces the inference speed. The second reason is that maintaining high-resolution image feature computation throughout the network consumes more computational resources. Nevertheless, our proposed model still holds advantages over CNN-based models, managing to achieve superior mIoU on difficult datasets with a relatively small increase in complexity. We provide visual comparison results with other methods in Figure \ref{fig:loveda}.

\subsubsection{Potsdam}
As a widely-used dataset for segmentation tasks, Potsdam can comprehensively demonstrates the improvement of the accuracy of HRS images by the model proposed in this paper. Table \ref{table:11} shows the scores achieved on the Potsdam dataset. The FLOPs and FPS are measured by 512$\times$512 inputs on a single NVIDIA GTX 3090 GPU. The network proposed in this paper achieves the F1 of 92.4\% and mIoU of 87.6\% on the Potsdam dataset. Hi-ResNet outperforms the lightweight convolutional network FANet~\cite{hu2020real} and the lightweight transformer-based network Segmenter~\cite{strudel2021segmenter}. Notably, Hi-ResNet performs well among all methods for the ``Lowveg'' class, achieving a score of 87.9\%. The car category also achieved a high score with a mean F1 of 96.1\%. This result fully demonstrates that the Hi-ResNet has better performance for small target segmentation in HRS images.
 
Moreover, when the input image resolution is reduced from 1024$\times$1024 to 512$\times$512, the complexity of Hi-ResNet significantly decreases, and the model's inference speed accelerates nearly sixfold. At this point, our proposed model can achieve faster inference speeds than Transformer-based networks such as RSSFormer~\cite{xu2023rssformer} and DC-Swin~\cite{wang2022novel}, even under conditions of higher complexity. This validates the performance advantage of our network.

We present Potsdam segmentation results to showcase the effectiveness of Hi-ResNet for small object segmentation. As displayed in Figure \ref{fig:potsdam}, most networks perform poorly in the segmentation of object edges. To overcome this limitation, Hi-ResNet employs CEA loss in the loss calculation to maximize the distance between the two boundaries, ensuring good connectivity of the extracted edge features during loss calculation, thus avoiding category boundary blur.

\subsubsection{Vaihingen}
The Vaihingen dataset has a large number of houses obscured by tree branches and multi-story small villages, so the dataset requires the network to identify and segment small targets more accurately. 
Table \ref{table:12} shows the results of different methods on the Vaihingen dataset. (Note that the calculation for FLOPs and FPS is the same as for Potsdam). Our proposed network achieves an OA of 91.7\% and the mIoU of 79.8\% on the Vaihingen dataset. In the low vegetation category, Hi-ResNet secured first place with the same performance on the Potsdam dataset. Significantly, our network improves the results for the categories of ``Building'' and ``Car'' by 4\% compared to the HRNet network. This is because Hi-ResNet effectively solves the sample imbalance problem caused by small targets occupying small pixels in HRS images by using the CEA loss and GD loss to weigh each category.

We show some typical segmentation results in vaihingen in Figure \ref{fig:vaihingen}. Most networks present misclassification on the ``Tree'' class and the ``Lowveg'' class. At the same time, the within-class distance segmentation redundancy of the dense small target object ``Car'' class, and the edge segmentation of the small cars is not clear. Hi-ResNet can obtain the position and edge information of small cars more accurately when the global receptive field is increased, thereby avoiding misclassification in complex scenes. 

\section{Conclusion}
Our study centers on the semantic segmentation of HRS, specifically focusing on addressing the inherent challenges of object scale and shape variance, and complex background environments. These issues often lead to object misclassification and sub-optimal outcomes with current learning algorithms. We respond by developing Hi-ResNet, which stands out due to an efficient network structure that includes a funnel module, a multi-branch module embedded with IA blocks, and a feature refinement module. Additionally, we introduce the CEA loss function. In our approach, the funnel module functions to downsample and extract high-resolution semantic information from the input image. The process then moves to the multi-branch module with stacks of IA blocks, enabling the capture of image features at different scales and distinguishing variant scales and shapes within the same class. Our study concludes with the integration of the CEA loss function within our feature refinement module. This innovative step effectively disambiguates inter-class objects with similar shapes and increases the data distribution distance for accurate predictions. The superiority of Hi-ResNet is proven through a comparative evaluation with leading methodologies across LoveDA benchmarks. The results underscore the value of our contributions to advancing HRS semantic segmentation and demonstrate the sensitivity of parallel architecture of the input size.

\bibliographystyle{IEEEtran}
\bibliography{reference}
\end{document}